\documentclass{article}

\PassOptionsToPackage{numbers,sort&compress}{natbib}
\usepackage[preprint]{neurips_2026}

\usepackage[utf8]{inputenc} 
\usepackage[T1]{fontenc}    
\usepackage{hyperref}       
\usepackage{url}            
\usepackage{booktabs}       
\usepackage{amsfonts}       
\usepackage{nicefrac}       
\usepackage{microtype}      
\usepackage{xcolor}         
    
\usepackage{float}
\usepackage{stfloats} 
\usepackage{graphicx}
\usepackage{algorithm}
\usepackage[noend]{algpseudocode}
\usepackage{amsthm,amsmath, amssymb}
\usepackage{booktabs, ctable}
\usepackage{multirow}
\usepackage{enumitem}
\usepackage{bm}

\title{Hierarchical Scaffolding Enables Human-Like Cognitive Selectivity under Data Scarcity}

\author{
  Juhyoung Park$^{1}$ \quad Jaehyuk Bae$^{2}$ \quad Hyeonbo Yang$^{2}$ \quad Se-Bum Paik$^{2}$\\
  $^1$School of Computing \\
  $^2$Department of Brain and Cognitive Sciences \\
  Korea Advanced Institute of Science and Technology, Daejeon, Republic of Korea \\
  \texttt{\{juhpark, jaehyukbae, hyeonbo, sbpaik\}@kaist.ac.kr}\\
}

\begin{document}

\maketitle

\begin{abstract}
    Modern machine learning systems demand extensive datasets for visual recognition. Conversely, humans learn with high efficiency despite severe data limitations, often by acquiring broad categorical structures before refining finer distinctions. Inspired by this contrast, we introduce SCALA (Scaffolded Cognitive Architecture for Learning under limited dAta), a hierarchical learning framework grounded in cognitive psychology that guides models from coarse conceptual structures to fine-grained recognition. Our model exhibits human-like cognitive selectivity by effectively prioritizing task-relevant features while suppressing background distractors, a mechanism that induces a fundamental shift in representation learning. This shift is characterized by accelerated cluster formation, reduced intra-class dispersion, and enhanced semantic separability. Empirically, SCALA achieves significant accuracy improvements under severe data scarcity. Furthermore, this hierarchical scaffolding promotes robust generalization to unseen classes and accelerates the acquisition of novel categories. Collectively, our results establish SCALA as a powerful framework for achieving human-level sample efficiency and resilient category generalization in data-constrained environments.  
\end{abstract}

\section{Introduction}

Recent advances in deep learning have shown that performance can improve by scaling data, model size, and computation, often following empirical scaling laws \cite{ref01}. Large-scale models further illustrate how strong performance can be achieved through massive data and compute \cite{ref02,ref03}. However, this strategy comes with growing resource demands that cannot always be met in practice. In many real-world settings, datasets are limited, imbalanced, or not fully representative of deployment conditions \cite{ref04,ref05}. This is especially visible in specialized domains where data collection is expensive, slow, or inherently constrained; one clear example is rare-disease research, where low prevalence and clinical heterogeneity can make it difficult to assemble datasets of sufficient size for deep learning \cite{ref06}. Under such conditions, scaling alone is not a viable solution. 

\begin{figure}[t]
	\centering
	\includegraphics[width=\textwidth]{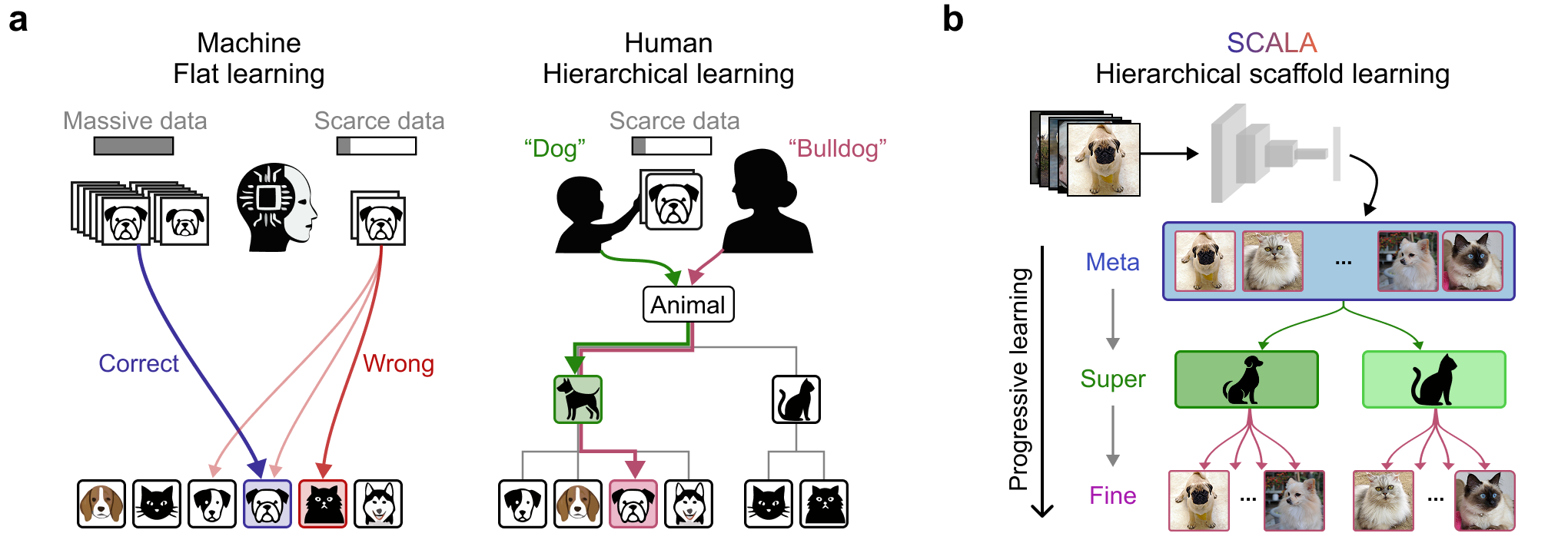}
    \vspace{-0.5cm}
    \caption{\textbf{Schematic framework of our proposed method: SCALA.}
        (a) Conventional models learn directly from flat labels, which can lead to poor performance under data scarcity. In contrast, humans acquire hierarchical concepts, enabling more sample-efficient learning and robust categorization.
        (b) Hierarchical scaffold learning restructures the label space into a semantic hierarchy. Early coarse-level supervision provides a structural scaffold that guides progressive refinement of fine-grained distinctions, resulting in improved learning efficiency and representation quality.
    }
	\label{fig1}
\end{figure}

Human learners provide a striking contrast (Fig.~\ref{fig1}a). Children can generalize the meaning of a novel word from one or a few positive examples \cite{ref07}, two-year-olds can initiate category formation from only a small number of labeled exemplars and then incorporate later unlabeled exemplars into the same emerging category representation \cite{ref08}, and adults can infer novel visual concepts from a single exemplar \cite{ref09}. These findings suggest a mode of category learning that is far more sample-efficient than the large-data regime that typically underwrites strong machine-learning performance. The same representational economy matters when learned categories are used under ambiguity: a learner must not only acquire a category from limited evidence, but also preserve the relevant category when other visual evidence competes for the decision.

\setlength{\textfloatsep}{8pt}
\setlength{\floatsep}{8pt}
\setlength{\intextsep}{8pt}

A familiar case is visual CAPTCHA: selecting all images that contain a queried object category. When asked to “select all images containing a bicycle,” humans can often identify the bicycle even when it appears beside a person, a car, a road sign, or other salient objects. For machine systems, the difficulty is not simply to summarize an image with a global label, but to decide whether the queried object is present while other salient objects compete for the decision. Common engineering routes—more data, specialized attention mechanisms, or more elaborate inference-time procedures—can improve performance, but they shift the burden to data, architecture, and computation. This motivates the question of whether part of this ability can be encouraged through the structure of supervision itself.

In cognitive science, this efficiency has long been associated with structured conceptual organization rather than flat sets of isolated labels. Categories span multiple abstraction levels, such as superordinate, basic, and subordinate levels \cite{ref10} and learning proceeds by integrating new information into existing structures while progressively differentiating into more specific representations \cite{ref11,ref12}, with intermediate support scaffolding this refinement \cite{ref13,ref14}. These perspectives suggest that coarse distinctions are learned before fine-grained ones, supporting efficient category acquisition.

In machine learning, however, standard supervised learning remains structurally "flat": labels are treated as independent categories, ignoring relationships among classes. While abundant data may allow models to recover semantic organization \cite{ref34,ref35}, flat supervision overloads limited samples under data scarcity: requiring models to simultaneously learn fine distinctions and broader semantic organization. This raises a central question: can explicitly structuring the label space improve sample efficiency and generalization when fine-grained supervision is scarce? The answer may lie in hierarchical priors underlying biological learning that enable rapid knowledge acquisition.

Such perspectives on human learning have inspired machine learning approaches that incorporate hierarchy into visual recognition through multi-level prediction \cite{ref18}, coarse-to-fine curricula over the output space \cite{ref15}, basic-level pretraining and hierarchical linguistic supervision \cite{ref16,ref17}, as well as hierarchy-aware objectives and architectures \cite{ref21,ref22,ref23,ref24}. Despite these advances, these approaches typically treat hierarchy as an auxiliary constraint or rely on model-derived hierarchies whose semantic reliability may be limited under data scarcity. In contrast, cognitive accounts suggest that biological learning is guided by hierarchical priors, in which coarse conceptual distinctions are acquired before refinement into specific representations.

Motivated by this perspective, we propose a framework that treats a hierarchy as an explicit scaffold for learning rather than an  auxiliary objective or output structure. Instead of treating fine-grained classification as a flat prediction problem over independent labels, we restructure the label space into a semantic hierarchy grounded in principles of human category organization that guides the model through a trajectory of progressive differentiation (Fig.~\ref{fig1}b). This hierarchical scaffolding keeps a single classifier throughout training and derives coarse-level objectives by aggregating logits, thereby organizing the learning trajectory without introducing hierarchy-specific heads.

\noindent\textbf{Our main contributions are as follows:}
\begin{itemize}[noitemsep, topsep=0pt, leftmargin=16pt]
    \item \textbf{Human-like selective attention}: Our framework prioritizes task-relevant features while suppressing distractors, inducing a fundamental shift in representation learning that recapitulates key hallmarks of human cognitive selectivity (Sec.~\ref{sec3.1}).
    \item \textbf{Sample-efficient learning}: Hierarchical scaffolding significantly improves learning performance over conventional flat-learning under severe data scarcity (Sec.~\ref{sec3.1}).
    \item \textbf{Enhanced representation learning}: By learning coarse semantic structure before fine-grained distinctions, our model produces more compact and separable representations across the hierarchy while progressively organizing representations toward full-data-like structures during learning (Secs.~\ref{sec3.2}--\ref{sec3.3}).
    \item \textbf{Generalization to unseen classes}: Our framework facilitates zero-shot generalization to unseen classes at higher semantic levels and supports efficient adaptation to novel categories (Sec.~\ref{sec3.4}).
\end{itemize}

\section{Hierarchical scaffolding framework}

\subsection{Hierarchical scaffolding}

We introduce SCALA (Scaffolded Cognitive Architecture for Learning under limited dAta), a hierarchical scaffolding framework that keeps a single fine-class classifier throughout training while deriving coarser semantic predictions by aggregating fine-class logits.

Let $M$ denote the number of hierarchy levels, indexed from coarse to fine. $\mathcal{Y}^{(1)}$ is the coarsest label space, and $\mathcal{Y}^{(M)}$ is the finest label space with $C$ classes. Each fine class in $\mathcal{Y}^{(M)}$ is assigned to exactly one group at every coarser level, forming a nested hierarchy, so all coarser labels are uniquely determined by the fine label. In our main CIFAR-100 experiments~\cite{krizhevsky2009learning}, we use a three-level hierarchy ($M=3$): 100 fine classes, 20 dataset-provided superclasses, and five meta-classes formed by merging semantically related superclasses (Table \ref{tab:cifar100_hierarchy}).

Given an input $x$, the network outputs logits only over the $C$ fine classes,
\begin{equation}
z(x) = f_\theta(x) \in \mathbb{R}^C.
\end{equation}
Rather than introducing separate classification heads for coarser levels, SCALA derives coarser predictions from fine-level logits. Let $G_k^{(m)} \subseteq \{1,\dots,C\}$ denote the set of fine classes assigned to group $k$ at level $m$, where $m \in \{1,\dots,M-1\}$. The corresponding group logit is defined as
\begin{equation}
z_k^{(m)}(x)=\log\sum_{c\in G_k^{(m)}} \exp(z_c(x)).
\end{equation}

Because the groups at each level partition the fine-class label space, this aggregation makes coarse-level prediction consistent with the fine classifier: after softmax normalization, each group probability equals the sum of the probabilities of its member fine classes. Thus, SCALA introduces hierarchy through label organization and logit aggregation, while leaving the classifier architecture unchanged.

SCALA then trains the model progressively from coarse to fine. At stage $m$, each fine label $y_i^{(M)}$ is mapped to its corresponding level-$m$ label,
\begin{equation}
y_i^{(m)} = g^{(m)}(y_i^{(M)}),
\end{equation}
and the model is optimized with a standard cross-entropy loss at that level,
\begin{equation}
\mathcal{L}^{(m)}
=
\frac{1}{|\mathcal{B}|}
\sum_{i \in \mathcal{B}}
\mathrm{CE}\left(z^{(m)}(x_i), y_i^{(m)}\right),
\end{equation}
where $z^{(m)}(x_i)$ denotes the hierarchy-aggregated logits for $m<M$, and $z^{(M)}(x_i)=z(x_i)$ denotes the original fine-class logits.

The loss is applied sequentially across hierarchy levels. In our main three-level setting, SCALA first optimizes the coarsest 5-way meta-class objective, then the 20-way superclass objective, and finally the 100-way fine-class objective. This schedule encourages the model to form broad semantic neighborhoods before refining them into fine-grained class boundaries, without introducing additional classifier heads or changing the final prediction task.

\subsection{Cognitive grounding of the CIFAR-100 label hierarchy}
\label{sec2.2}

To assess whether the hierarchy used in our experiments reflects human cognition, we conducted a behavioral grouping task with 9 participants. Participants assigned fine classes to fixed superclass labels and then assigned superclasses to fixed meta-class labels.

Human assignments closely matched the predefined hierarchy at both levels, with 94.78\% $\pm$ 4.55\% agreement for superclasses and 92.67\% $\pm$ 4.80\% agreement for meta-classes. Pairwise agreement was also high at both levels (mean $\textit{phi}$ = 0.906 $\pm$ 0.084 for superclasses; 0.815 $\pm$ 0.118 for meta-classes). These results indicate that the hierarchy used in this study is strongly aligned with human cognition. Additional procedural details are provided in Appendix~\ref{app:hier}.

\section{Results}
\subsection{Hierarchical scaffolding promotes human-like attention under severe data scarcity}
\label{sec3.1}

\begin{figure}[h!]
	\centering
	\includegraphics[width=\textwidth]{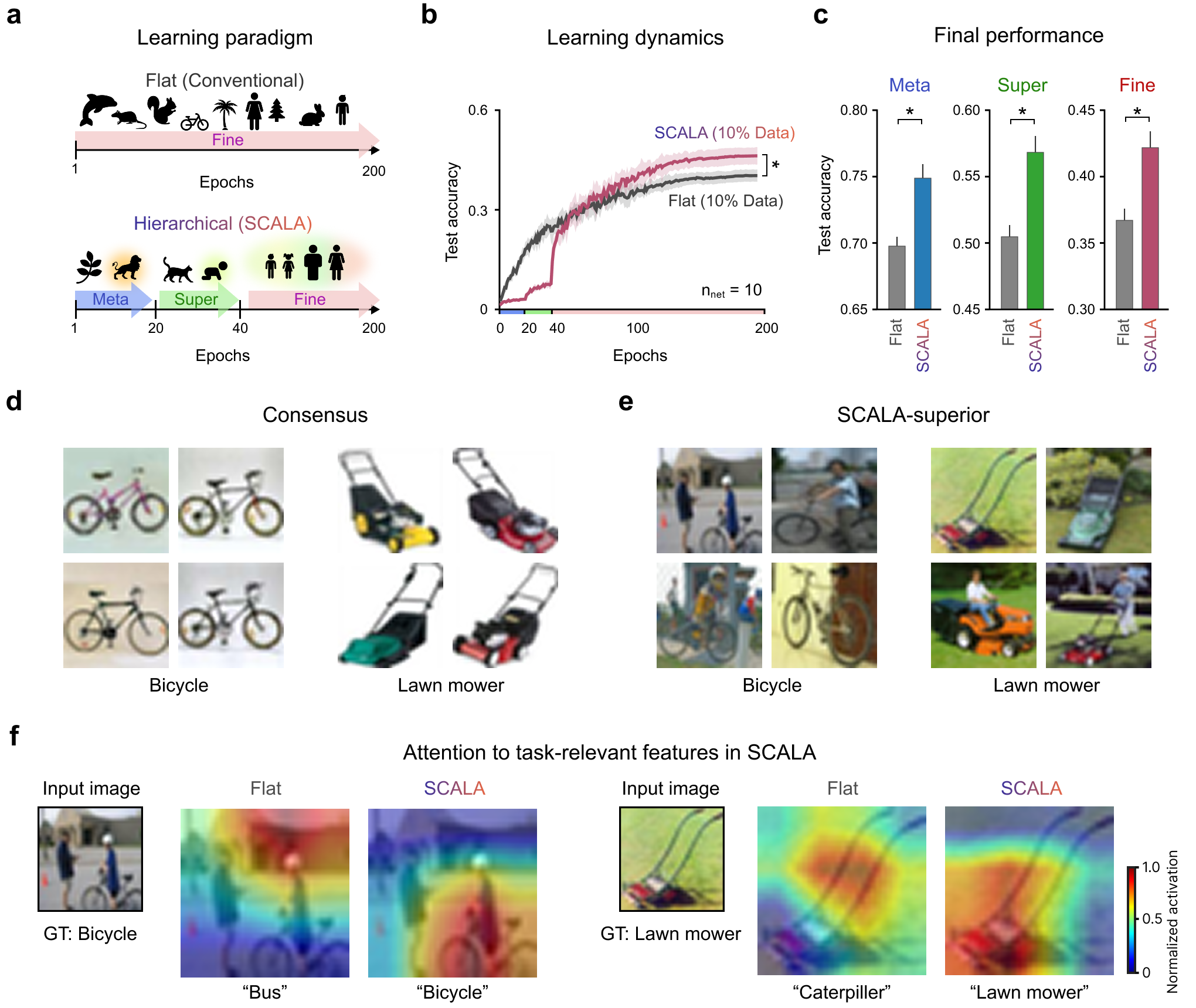}
    \vspace{-0.5cm}
    \caption{\textbf{Hierarchical scaffolding enhances performance via human-like selective attention under data scarcity.}
    (a) A conceptual overview contrasting conventional flat-learning with hierarchical learning of SCALA.
    (b) Test accuracy over training epochs under a 10\% data-limited condition.
    (c) Final test accuracy at the meta-, super-, and fine-grained levels for flat-learning and SCALA.
    (d) Representative consensus samples correctly classified by both flat-learning and SCALA across 10 independent runs.
    (e) Challenging samples where SCALA showed consistently superior performance in predicting the correct label while flat-learning failed, often involving significant distractors, spurious background correlations, or multiple objects.
    (f) Grad-CAM visualization for SCALA-superior samples, demonstrating that while flat-learning attends to background-correlated or irrelevant objects, SCALA focuses consistently on task-relevant features. Text below each heatmap indicates the model prediction; GT denotes the ground-truth class. Colorbar indicates normalized Grad-CAM intensity.
    }
	\label{fig2}
\end{figure}

To evaluate whether hierarchical scaffolding improves learning under data-limited conditions, we compare SCALA against a conventional flat-learning model using a ResNet-18 architecture~\cite{he2016deep} trained on a 5k subset (10\%) of the CIFAR-100 dataset  (Fig.~\ref{fig2}a). For a fair comparison, both SCALA and the flat-learning model are trained for 200 epochs using identical optimization settings. Unless otherwise noted, all subsequent analyses are based on this 5k training subset.

While SCALA was trained with higher-level objectives in the early training stage, its accuracy increased rapidly once fine-grained supervision was introduced (Fig.~\ref{fig2}b). Despite delaying direct fine-class supervision, SCALA does not lag behind the flat-learning model; instead, the initial hierarchical stages appear to cultivate a representational foundation where fine-grained boundaries can be learned more efficiently. Consequently, SCALA outperforms the flat-learning counterpart, achieving significantly higher final test accuracy across meta-, super-, and fine-level evaluations (Fig.~\ref{fig2}c; Flat vs. SCALA, $n_{\text{net}}=10$, Wilcoxon rank-sum test, $\textit{p}<0.05$) and reaches the conventional model’s best accuracy much earlier (98.3 vs. 183.4 epochs). Similar test accuracy improvements generalize across architectures and datasets under data scarcity (Appendix~\ref{app:generalization_datasets_architectures}).

To investigate the underlying mechanisms of these performance gains, we examined the attentional focus of each model during inference. We analyzed two distinct categories of samples: (i) consensus samples, correctly classified by both models, and (ii) SCALA-superior samples, correctly classified only by our framework. Model attention was visualized using Grad-CAM~\cite{ref44}. For "consensus" samples, images typically feature canonical object views with clean backgrounds and a single, centered task-relevant object (Fig.~\ref{fig2}d). In contrast, "SCALA-superior" samples often involve distracting or ambiguous visual cues, such as background clutter, which interfere with the flat-learning model's ability to identify the target object (Fig.~\ref{fig2}e). Remarkably, Grad-CAM visualizations reveal that SCALA consistently prioritizes object-relevant regions, whereas the flat-learning counterpart often attends to localized or spurious background-correlated cues (Fig.~\ref{fig2}f). These patterns suggest that hierarchical scaffolding encourages the model to prioritize task-relevant features while mitigating reliance on spurious correlations, even under severe data scarcity. These observations were consistent across a larger set of samples (Appendix~\ref{app:attention_analysis}).

This selective emphasis mirrors human perceptual strategies, where task-relevant features are preferentially processed in the presence of ambiguity \cite{cheon2025}. By internalizing coarse concepts first, SCALA introduces an inductive bias that guides the representation toward semantically grounded features, enabling the model to focus on the target object rather than competing distractors. Supporting this interpretation, a human image-classification task showed that images selectively classified correctly by SCALA, but less reliably by flat-learning, were easily categorized by human observers (96.8\% accuracy). Thus, SCALA’s advantage was expressed on images that were perceptually clear to humans but not consistently captured by flat-learning, suggesting that SCALA closely reflects human-like visual categorization (Appendix~\ref{Human perceptual validation}). Furthermore, we observed that even when both models err, SCALA’s predictions tend to remain within the correct superclass, whereas flat-learning often yields semantically irrelevant misclassifications; conversely, when both are correct, SCALA assigns higher confidence to the correct class (Appendix~\ref{app:both_wrong_correct}).

\subsection{Hierarchical scaffolding shapes semantically structured representations}
\label{sec3.2}

\begin{figure}[h!]
	\centering
	\includegraphics[width=\textwidth]{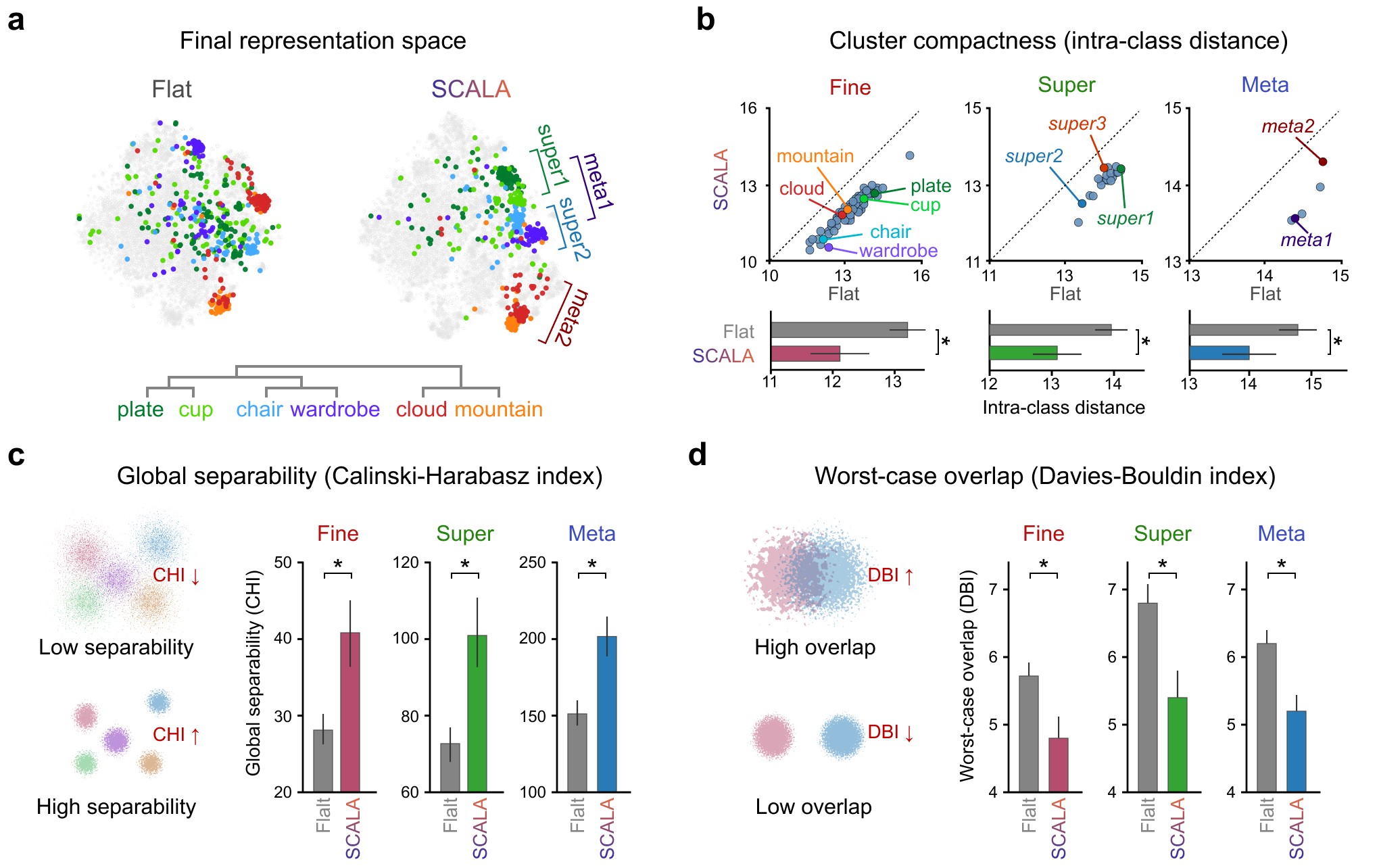}
    \vspace{-0.5cm}
    \caption{\textbf{Hierarchical alignment and compactness of representation geometry.}
    (a) Final penultimate-layer embeddings visualized via t-SNE for the flat-learning model and SCALA.
    Highlighted classes are drawn from two meta-classes in the semantic hierarchy. Meta1
    (\textit{Indoor\_objects}) includes \textit{food\_containers}: \textit{plate}, \textit{cup} and
    \textit{household\_furniture}: \textit{chair}, \textit{wardrobe}, whereas Meta2
    (\textit{Outdoor\_objects\_and\_scenes}) includes \textit{large\_natural\_outdoor\_scenes}:
    \textit{cloud}, \textit{mountain}.
    (b) Structural compactness quantified by intra-class distance at the meta-, super-, and fine-levels. SCALA consistently reduces intra-class distance, reflecting tighter cluster formation.
    (c) Global separability evaluated through the Calinski--Harabasz index (CHI).
    (d) Worst-case cluster overlap assessed by the Davies--Bouldin index (DBI). All comparisons (Flat vs. SCALA) are statistically significant ($n_{\text{net}}=10$, Wilcoxon rank-sum test, $\textit{p}<0.05$).
    }
	\label{fig3}
\end{figure}

Next, we examined how hierarchical scaffolding reshapes the geometry of the final representation space. To visualize the development of this space under data-limited conditions, we applied t-SNE~\cite{maaten2008} to the last penultimate-layer embeddings. While flat-learning produced diffuse and partially overlapping class organization, with highlighted categories dispersed across the feature space, SCALA generated a more semantically structured geometry (Fig.~\ref{fig3}a). In SCALA’s embedding, fine-grained classes formed compact clusters, which were further organized into their respective superclasses and broader meta-class structures. This suggests that hierarchical scaffolding does not merely improve fine-class discrimination, but it encourages representations to inherently reflect the semantic hierarchy used during training.

We quantified these structural differences across the meta-, super-, and fine-levels using intra-class distance, the Calinski--Harabasz index (CHI)~\cite{calinski1974dendrite}, and the Davies--Bouldin index (DBI)~\cite{davies1979cluster}. Compared to flat-learning, SCALA consistently reduced intra-class distance across all three hierarchical levels (Fig.~\ref{fig3}b; Flat vs. SCALA, $n_{\text{net}}=10$, Wilcoxon rank-sum test, $\textit{p}<0.05$), indicating that samples from the same category were drawn into tighter, more cohesive clusters. Notably, this improvement was evident even at the fine-class level, despite SCALA bypassing direct fine-class optimization during the initial 40 epochs in favor of higher-level supervision.

Furthermore, SCALA significantly enhanced cluster separation. It yielded higher CHI values across all levels (Fig.~\ref{fig3}c; Flat vs. SCALA, $n_{\text{net}}=10$, Wilcoxon rank-sum test, $\textit{p}<0.05$), reflecting superior global separability. Simultaneously, SCALA achieved lower DBI values (Fig.~\ref{fig3}d; Flat vs. SCALA, $n_{\text{net}}=10$, Wilcoxon rank-sum test, $\textit{p}<0.05$), indicating a marked reduction in overlap between neighboring clusters. Together, these results show that hierarchical scaffolding produces more compact, semantically coherent, and well-separated robust representations under severe data scarcity.

\subsection{Early coarse supervision aligns representations toward full-data-trained structures}
\label{sec3.3}

\begin{figure}[h!]
	\centering
	\includegraphics[width=\textwidth]{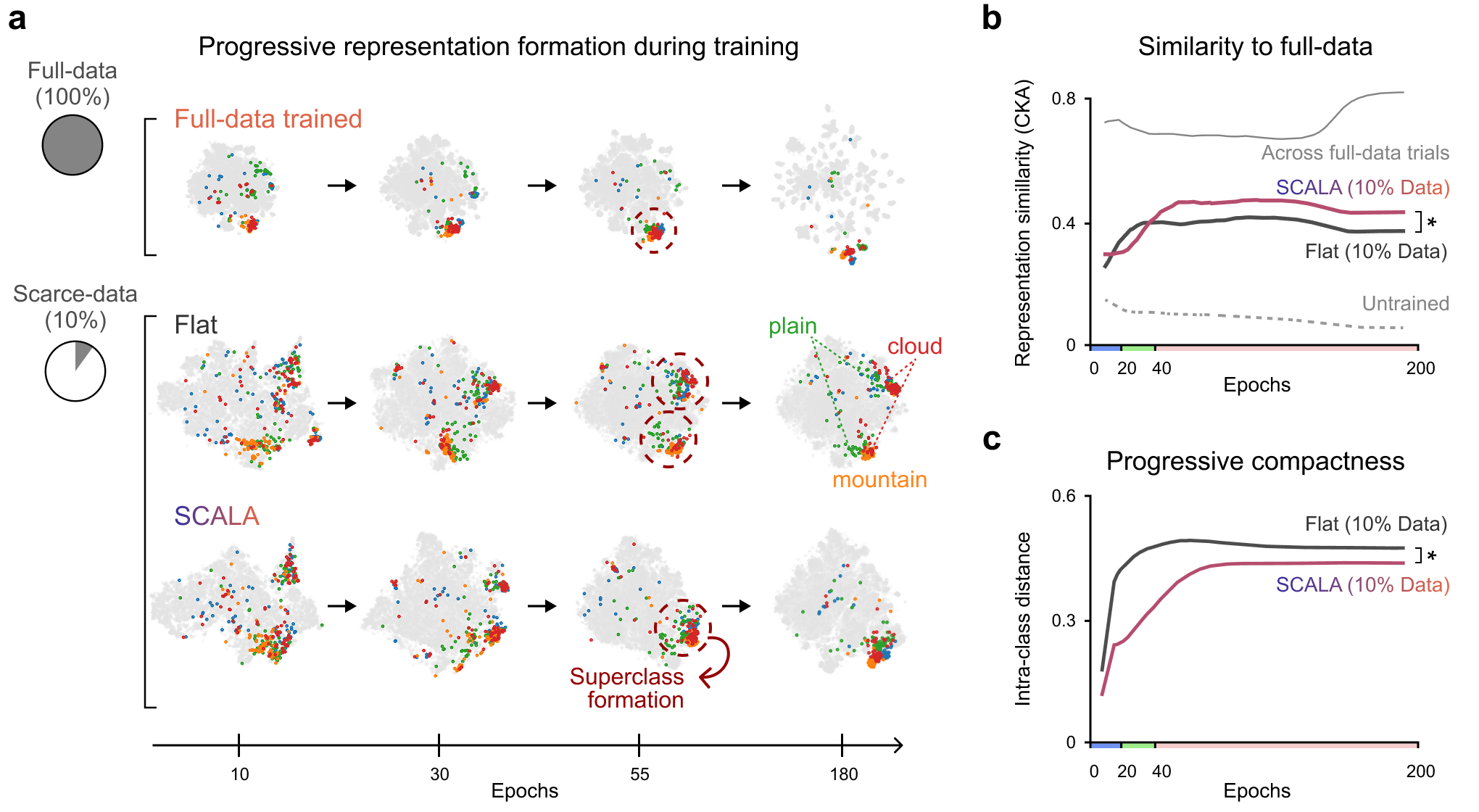}
    \vspace{-0.5cm}
    \caption{\textbf{Progressive representation geometry analysis.}
    (a) Evolution of representation geometry visualized with t-SNE in a 10\% data regime. The flat-learning model and SCALA are compared across epochs 10, 30, 55, and 180, with the full-data model provided as a reference. Classes within \textit{large\_natural\_outdoor\_scenes} superclass are highlighted: \textit{cloud} (red), \textit{mountain} (orange), \textit{plain} (green), and \textit{sea} (blue).
    (b) Representational similarity to the full-data reference at the final residual stage of ResNet-18. The gray dashed line indicates an untrained network, while the gray solid line denotes the similarity between independently trained full-data trials.
    (c) Fine-level intra-class cosine distance across training. The gray solid line denotes the full-data trained model.
    }
    
	\label{fig4}
\end{figure}

To investigate how early-stage coarse scaffolds guide the development of internal representations, we applied t-SNE to penultimate-layer features across training epochs (Fig.~\ref{fig4}a). In the 10\% data regime, flat-learning model failed to reproduce the organized neighborhood structure observed in the full-data reference; for instance, related classes within \textit{large\_natural\_outdoor\_scenes} remained dispersed throughout the embedding space rather than forming a coherent semantic region. In contrast, SCALA maintained these classes within a compact neighborhood that closely mirrored the geometry of the full-data model. This effect emerged progressively; the highlighted classes first condensed into a shared semantic region before bifurcating into distinct, fine-grained clusters. This suggests that SCALA follows a representation formation trajectory that effectively recapitulates the structural development seen in models trained with abundant data.

We further quantified this structural alignment by measuring representational similarity between each 10\%-data model and the full-data reference using Centered Kernel Alignment (CKA)~\cite{ref27} at the final residual stage of ResNet-18. SCALA exhibited significantly higher similarity to the full-data reference than flat-learning (Fig.~\ref{fig4}b; Flat vs. SCALA, $n_{\text{net}}=10$, Wilcoxon rank-sum test, $\textit{p}<0.05$). This result supports the visual pattern in Fig.~\ref{fig4}a, confirming that hierarchical scaffolding induces a more full-data-like representational structure throughout the network under severe data limitation.

Additionally, we examined directional alignment of representations by tracking intra-class cosine distance across training. Both models exhibited an initial increase in intra-class distance, reflecting early representation disentanglement from their initially unstructured state. However, SCALA showed a markedly stronger subsequent compression, particularly during the superclass stage (epochs 20–40), where coarse supervision grouped semantically related classes into shared neighborhoods. This early consolidation persisted after transitioning to fine-level supervision, yielding consistently tighter clusters than flat-learning. The gap established at intermediate stages was preserved through convergence, resulting in significantly lower final intra-class distance (Fig.~\ref{fig4}c;  Flat vs. SCALA, $n_{\text{net}}=10$, Wilcoxon rank-sum test, $\textit{p}<0.05$). These results suggest that early coarse supervision acts as a semantic boundary that organizes fine-level representations, promoting progressively compact and well-separated structures that recapitulate full-data geometry under data scarcity.


\subsection{Generalization and adaptation to unseen classes from structured representations}
\label{sec3.4}

\begin{figure}[h!]
    \centering
    \includegraphics[width=\textwidth]{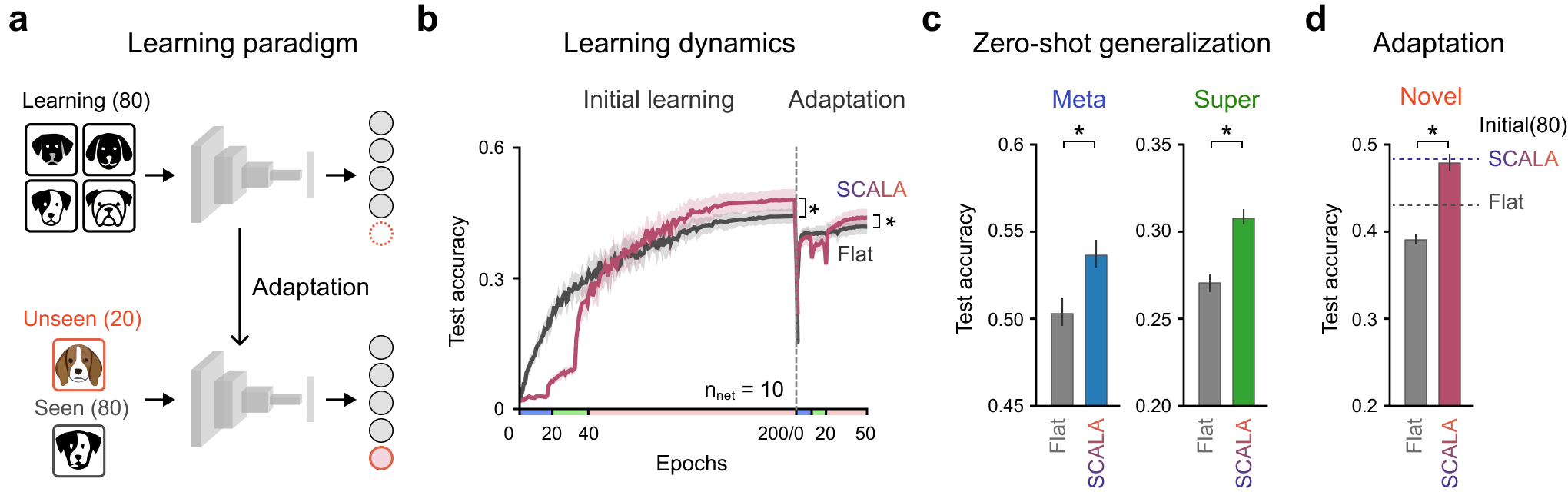}
    \vspace{-0.5cm}
    \caption{\textbf{Generalization and adaptation to unseen classes.}
    (a) Experimental setup in which models are first trained on 80 seen classes and subsequently adapted to all 100 classes after reintroducing the 20 unseen classes.
    (b) Test accuracy over training with an initial phase (0–200 epochs) on the 80 seen classes followed by an adaptation phase (0–50 epochs) on all classes.
    (c) Zero-shot higher-level generalization after initial 80 class learning, where images from the 20 unseen fine classes are correctly assigned to their corresponding meta- and super-level categories without additional training.
    (d) Fine-class test accuracy on the 20 unseen classes after additional training with all 100 classes, where dashed lines indicate the final test accuracy of initial 80 class learning.}
    \label{fig5}
\end{figure}

Building on the observation that hierarchical scaffolding yields highly structured representations, we next investigated whether this organization facilitates the integration of novel categories. In cognitive development, new knowledge is not acquired in isolation but is integrated into an existing, continuously refined conceptual framework \cite{ref10, ref11, ref12, ref13, ref14}. To examine this adaptability, we designed an unseen-class experiment using CIFAR-100 (Fig.~\ref{fig5}a). We held out one fine-level class from each superclass, resulting in 80 seen and 20 unseen categories. Both the flat-learning model and SCALA were trained initially on the 80 seen classes while maintaining a 100-way classifier head, with weights for unseen classes left untrained.

We then assessed two key capabilities: (i) whether the learned representation could accurately categorize unseen-class images into their correct higher-level groups without explicit training, and (ii) whether the model could quickly learn new, previously unseen classes once they were introduced. After training on the 80 classes, SCALA exhibited a rapid increase upon fine-grained supervision and achieved significantly higher final accuracy than the flat-learning counterpart (Fig.~\ref{fig5}b; Flat vs. SCALA, $n_{\text{net}}=10$, Wilcoxon rank-sum test, $\textit{p}<0.05$). At this stage, prior to adaptation, we evaluated zero-shot higher-level generalization using test images from the 20 unseen fine classes. Despite never having encountered these specific categories, SCALA assigned them to the correct meta and superclasses significantly more accurately than the flat-learning counterpart (Fig.~\ref{fig5}c; Flat vs. SCALA, $n_{\text{net}}=10$, Wilcoxon rank-sum test, $\textit{p}<0.05$). This indicates that hierarchical scaffolding organizes the representation space in a manner that transcends specific observed classes, allowing novel categories to be embedded within a robust, pre-existing semantic structure. This advantage was particularly pronounced under severe data scarcity (Appendix~\ref{app:unseen_class_results}).

Next, we introduced the 20 unseen classes and continued training on all 100 classes for 50 epochs. During this adaptation phase, the flat-learning model continued standard fine-label training, whereas SCALA applied the same hierarchical staged supervision as in initial training to the 20 unseen classes, while maintaining fine-level supervision for the 80 seen classes. This ensured that SCALA consistently refined its internal representation under a unified hierarchical principle.

Results demonstrate that SCALA significantly enhanced accuracy on the 20 novel categories compared to the flat-learning model, reaching performance close to its pre-adaptation level, whereas the flat-learning counterpart remained both lower in absolute accuracy and below its own pre-adaptation performance (Fig.~\ref{fig5}d; Flat vs. SCALA, $n_{\text{net}}=10$, Wilcoxon rank-sum test, $\textit{p}<0.05$). The performance on previously seen classes was preserved in both models, with a slight decrease after adaptation (Table~\ref{tab:seen80_after_adaptation}; 10\% data, Flat vs. SCALA, $n_{\text{net}}=10$, Wilcoxon rank-sum test, $\textit{p}>0.05$). This effect was pronounced under data scarcity, especially in the 10\% data regime (Table~\ref{tab:unseen_after_adaptation}), where SCALA improved fine-class accuracy on novel categories by up to $+9.14\%$, with corresponding gains at the super-class ($+5.68\%$) and meta-class ($+4.03\%$) levels. These results suggest that hierarchical scaffolding is particularly effective when new categories must be learned from sparse examples, as the pre-established semantic framework provides a stable scaffold to guide adaptation.

\section{Discussion}

\textbf{The geometry of structured learning.} Hierarchical scaffolding offers a fundamental lesson for data-efficient recognition: the sequence of learning objectives dictates the utility of subsequent data. In SCALA, coarse labels are not merely "simpler" targets; they establish a semantic coordinate system that preserves neighborhood structures before fine-grained evidence becomes reliable. Under severe data scarcity, this ordering fundamentally reshapes the learning problem. Rather than constructing category organization from sparse, isolated points, fine-level supervision merely refines a pre-existing, robust structure into class-specific boundaries. This mechanism explains why representational gains transcend simple accuracy. Flat supervision forces sparse data to simultaneously define fine-class boundaries, separate them from all other categories, and induce a global semantic organization—a multi-task burden that often results in a fragile, fragmented geometry. SCALA avoids this by stabilizing broad partitions early and sharpening boundaries only within an already organized space. Our findings on CKA similarity, class compactness, and separability collectively point to a singular conclusion: SCALA transforms scarce examples from isolated labels into effective constraints on a structured representation.

\textbf{Convergence with cognitive and developmental theories.} Our results provide a computational instantiation of core developmental principles. Learning, in this framework, is not the accumulation of disconnected labels but the progressive differentiation of structured concepts. Rosch’s account of category abstraction suggests that concepts are organized across levels, from broad superordinate categories to more specific subordinate distinctions \cite{ref10}. Piagetian and Ausubelian accounts further emphasize that new information is interpreted through existing cognitive structures and gradually differentiated into more precise forms \cite{ref11, ref12}. Vygotskian and scaffolding theories add that intermediate support can make later independent performance possible by constraining what the learner must solve at each stage \cite{ref13, ref14}. SCALA operationalizes these ideas into a unified training principle. The contribution is not merely improved accuracy, but a developmentally inspired organization of supervision that fundamentally reshapes how machine representations become discriminative. 

\textbf{From selectivity to extensibility.} This representation-level effect provides a route toward cognitive selectivity without requiring dedicated attention modules. In a CAPTCHA prompt such as “select all images containing a bicycle,” the decisive challenge is not only knowing the class bicycle, but keeping the queried category stable while people, signs, roads, or other salient objects compete for the decision. Our Grad-CAM analysis suggests that the early stabilization of semantic neighborhoods renders target categories significantly less vulnerable to competing visual distractors. This implies that selectivity is not merely an emergent mechanism at inference time; rather, it is a structural consequence of how category representations were organized during the learning process. Furthermore, our unseen-class results extend SCALA’s relevance from efficient fitting to extensible learning. By producing a category space where novel concepts can be integrated as refinements of an existing scaffold rather than unrelated additions, SCALA moves toward classification systems that evolve naturally over time.

\textbf{Conclusion: hierarchy as a developmental scaffold.} Finally, our random-grouping control (Appendix~\ref{app:random}) clarifies that the benefit of SCALA lies not in the simplicity of the early task, but in its semantic alignment with the eventual fine-grained objectives. This elevates hierarchy from a mere auxiliary regularizer to a developmental scaffold for representation formation. By strategically ordering the structure of supervision, SCALA yields representations that are more sample-efficient, more adaptable, and more robust under competing evidence—paving a new path for brain-inspired artificial intelligence.

\section{Broader impacts and limitations}

\textbf{Broader impacts.}
Our findings offer a pragmatic framework for high-stakes, real-world applications where labeled data are scarce or prohibitively expensive to acquire \cite{ref04,ref30,ref05}. By demonstrating that learning efficiency depends on supervisory structure rather than mere data scale, we provide a pathway for robust recognition that minimizes the reliance on exhaustive fine-grained labeling. This "supervision-efficient" approach is particularly valuable for specialized domains—such as medical imaging or rare-species identification—where coarse-level expertise is more accessible. Furthermore, by establishing a semantic scaffold early, SCALA supports efficient integration of novel categories as structured refinements, rather than isolated additions. This structural adaptability suggests that hierarchical scaffolding can serve as a foundation for sustainable AI systems that evolve alongside expanding label spaces without the need for computationally expensive retraining from scratch.

\textbf{Limitations.}
While this study establishes the advantages of hierarchical scaffolding under limited supervision, certain constraints warrant further investigation. Our experiments utilized predefined, semantically coherent taxonomies; thus, the effectiveness may vary depending on the integrity and granularity of the underlying hierarchy. In scenarios where category relations are noisy, contested, or incompletely specified, the benefits of fixed scaffolding might be degraded. Future research should explore the robustness of this developmental principle under automatically induced or partially latent hierarchies, potentially bypassing the need for curated taxonomic metadata while preserving the advantages of staged structural learning.

\begin{ack}
This work was supported by the National Research Foundation of Korea (NRF) grants (RS-2022-NR070602 and RS-2025-02304581 to S.P.) and the Institute of Information \& Communications Technology Planning \& Evaluation (IITP) grants (RS-2026-25522672 and RS-2026-25518317 to S.P.).
\end{ack}

\medskip
{
\small
\bibliographystyle{unsrt}
\bibliography{references}
}


\appendix
\newpage
\setcounter{table}{0}
\setcounter{figure}{0}
\renewcommand{\thefigure}{\Alph{section}\arabic{figure}}
\renewcommand{\thetable}{\Alph{section}\arabic{table}}
\renewcommand{\theHfigure}{\Alph{section}\arabic{figure}}
\renewcommand{\theHtable}{\Alph{section}\arabic{table}}


\section{Experimental setup}
\label{app:setup}

\subsection{Algorithmic details of hierarchical scaffolding}
\label{app:hierarchical_scaffolding}

Algorithm~\ref{alg:scala_training} summarizes the stage-wise training procedure used by SCALA. 
Throughout training, the network maintains a single classifier over the finest label space. 
At each coarse stage, fine-class logits are aggregated according to the predefined hierarchy, and the resulting group logits are used with the corresponding coarse labels. 
At the final stage, the same classifier is trained directly with the original fine-class labels.

\begin{algorithm}[H]
\caption{Stage-wise training with hierarchical scaffolding}
\label{alg:scala_training}
\begin{algorithmic}[1]
\Statex \textbf{Input:} dataset $\mathcal{D}$, model $f_\theta$, hierarchy $\{\mathcal{Y}^{(m)}\}_{m=1}^M$, label mappings $\{g^{(m)}\}_{m=1}^M$, groups $\{G_k^{(m)}\}$, stage boundaries $0=\tau_0 < \tau_1 < \cdots < \tau_M = T$
\Statex \textbf{Note:} $g^{(M)}$ is the identity mapping and $z^{(M)}(x)=z(x)$.

\For{epoch $t = 1$ \textbf{to} $T$}
    \State $m \gets \min\{r : t \leq \tau_r\}$ \Comment{Select supervision level by stage boundary}
    \ForAll{mini-batches $(\mathbf{X}, \mathbf{Y}^{(M)}) \subset \mathcal{D}$}
        \State $\mathbf{Z} \gets f_\theta(\mathbf{X})$ \Comment{$\mathbf{Z}\in\mathbb{R}^{B\times C}$: fine-class logits}
        \State $\mathbf{Y}^{(m)} \gets g^{(m)}(\mathbf{Y}^{(M)})$ \Comment{Map fine labels to current hierarchy level}

        \If{$m < M$}
            \State Initialize $\mathbf{Z}^{(m)} \in \mathbb{R}^{B \times |\mathcal{Y}^{(m)}|}$
            \For{$k = 1$ \textbf{to} $|\mathcal{Y}^{(m)}|$}
                \State $\mathbf{Z}^{(m)}_{:,k} \gets \log \sum_{c \in G_k^{(m)}} \exp(\mathbf{Z}_{:,c})$ \Comment{Aggregate logits for group $k$}
            \EndFor
            \State $\mathcal{L} \gets \mathrm{CE}(\mathbf{Z}^{(m)}, \mathbf{Y}^{(m)})$ \Comment{Coarse-level loss}
        \Else
            \State $\mathcal{L} \gets \mathrm{CE}(\mathbf{Z}, \mathbf{Y}^{(M)})$ \Comment{Original fine-level loss}
        \EndIf

        \State Update $\theta$ using $\nabla_\theta \mathcal{L}$
    \EndFor
\EndFor
\end{algorithmic}
\end{algorithm}

\subsection{Evaluation at different hierarchy levels}
\label{app:hierarchical_eval}

We evaluate performance at the fine, superclass, and meta levels using the hierarchy defined in Table~\ref{tab:cifar100_hierarchy}. For fine-level evaluation, we use the original 100-way prediction. For meta- and super-level evaluation, we do not map the top-1 fine-class prediction to its parent group. Instead, we first aggregate the fine-class logits within each hierarchy group using log-sum-exp.

For group $k$ at level $m$, the grouped logit is defined as
\begin{equation}
z_k^{(m)}(x) = \log \sum_{c \in G_k^{(m)}} \exp(z_c(x)),
\end{equation}
where $G_k^{(m)}$ denotes the set of fine classes belonging to group $k$ at level $m$. The level-$m$ prediction is then obtained by
\begin{equation}
\hat{y}^{(m)} = \arg\max_k z_k^{(m)}(x),
\end{equation}
and compared with the corresponding ground-truth label
\begin{equation}
y^{(m)} = g^{(m)}(y^{(M)}),
\end{equation}
where $y^{(M)}$ is the original fine-class label. This aggregation is equivalent to selecting the hierarchy group with the largest total fine-class probability mass. The same evaluation procedure is applied to both the flat and SCALA models.

\subsection{Dataset and hierarchy construction}
\textbf{CIFAR-100.}
CIFAR-100 consists of 100 fine classes organized into 20 superclasses, with 5 fine classes in each superclass~\cite{krizhevsky2009learning}. We used the original CIFAR-100 training and test splits defined by the dataset. For data-limited experiments, we constructed 10\%, 20\%, and 40\% training subsets, corresponding to 5k, 10k, and 20k images, respectively, by randomly sampling from the original training set while preserving the original class balance. The 100\% setting used the complete 50k-image training set. We constructed an additional coarser level by grouping the 20 superclasses into five semantically related meta-classes (Table~\ref{tab:cifar100_hierarchy}).

\begin{table}[t]
\centering
\caption{CIFAR-100 hierarchy used in this study. The 100 fine classes are organized into 20 dataset-provided superclasses, which we further group into five semantic meta-classes.}
\label{tab:cifar100_hierarchy}
\small
\setlength{\tabcolsep}{4pt}
\renewcommand{\arraystretch}{1.12}
\begin{tabularx}{\textwidth}{
>{\raggedright\arraybackslash}p{0.16\textwidth}
>{\raggedright\arraybackslash}p{0.28\textwidth}
>{\raggedright\arraybackslash}p{0.48\textwidth}
}
\toprule
\textbf{Meta} & \textbf{Super} & \textbf{Fine} \\
\midrule

\multirow{5}{=}{Land mammals}
& people & baby, boy, girl, man, woman \\
& small mammals & hamster, mouse, rabbit, shrew, squirrel \\
& medium-sized mammals & fox, porcupine, possum, raccoon, skunk \\
& large omnivores and herbivores & camel, cattle, chimpanzee, elephant, kangaroo \\
& large carnivores & bear, leopard, lion, tiger, wolf \\
\midrule

\multirow{3}{=}{Plants}
& trees & maple, oak, palm, pine, willow \\
& flowers & orchids, poppies, roses, sunflowers, tulips \\
& fruit and vegetables & apples, mushrooms, oranges, pears, sweet peppers \\
\midrule

\multirow{5}{=}{Other creatures}
& aquatic mammals & beaver, dolphin, otter, seal, whale \\
& fish & aquarium fish, flatfish, ray, shark, trout \\
& reptiles & crocodile, dinosaur, lizard, snake, turtle \\
& non-insect invertebrates & crab, lobster, snail, spider, worm \\
& insects & bee, beetle, butterfly, caterpillar, cockroach \\
\midrule

\multirow{3}{=}{Indoor objects}
& food containers & bottles, bowls, cans, cups, plates \\
& household electrical devices & clock, computer keyboard, lamp, telephone, television \\
& household furniture & bed, chair, couch, table, wardrobe \\
\midrule

\multirow{4}{=}{Outdoor objects and scenes}
& large man-made outdoor things & bridge, castle, house, road, skyscraper \\
& large natural outdoor scenes & cloud, forest, mountain, plain, sea \\
& vehicles 1 & bicycle, bus, motorcycle, pickup truck, train \\
& vehicles 2 & lawn mower, rocket, streetcar, tank, tractor \\
\bottomrule
\end{tabularx}
\end{table}

\textbf{Unseen-class experimental protocol.}
For the unseen-class experiment in Sec.~\ref{sec3.4}, we held out one fine class from each superclass. The 20 held-out classes are \textit{otter}, \textit{flatfish}, \textit{butterfly}, \textit{crab}, \textit{turtle}, \textit{bowls}, \textit{lamp}, \textit{couch}, \textit{wolf}, \textit{cattle}, \textit{possum}, \textit{woman}, \textit{squirrel}, \textit{sunflowers}, \textit{apples}, \textit{pine tree}, \textit{house}, \textit{cloud}, \textit{pickup truck}, and \textit{rocket}.

\subsection{Architecture details}

For our main CIFAR-100 experiments, we use the standard ResNet-18 architecture from He et al.~\cite{he2016deep}. The only modification was the commonly used CIFAR stem: we replaced the original ImageNet stem with a $3 \times 3$ convolution with stride 1 and removed the initial max-pooling layer. We made no other architectural changes.

\subsection{Training details}

For our main CIFAR-100 experiments, all models were trained for 200 epochs using SGD. We used a batch size of 128, an initial learning rate of 0.1, momentum of 0.9, and weight decay of $5 \times 10^{-4}$. The learning rate followed a cosine annealing schedule over the full training run. The same optimization hyperparameters were used for both the flat-learning and SCALA models.

For the unseen-class adaptation experiment in Sec.~\ref{sec3.4}, we first trained the model on the 80 "seen" classes while keeping a single 100-way classifier head, masking the 20 "held-out" classes from both the loss and the softmax normalization. Starting from this 80-class checkpoint, we introduced the held-out 20 classes and continued training for 50 epochs. The flat-learning model used fine-level supervision for all 100 classes throughout adaptation, whereas SCALA kept fine-level supervision for the previously seen 80 classes and applied staged hierarchical supervision only to the newly introduced 20 classes: meta-level supervision for epochs 1--10, super-level supervision for epochs 11--20, and fine-level supervision for epochs 21--50.  The losses on the previously learned and reintroduced class subsets were weighted equally to balance learning between previously learned and newly introduced classes. The optimization settings were the same as in the main CIFAR-100 experiments, except that we used an initial learning rate of 0.01 with a cosine annealing scheduler starting at epoch 21.

\subsection{Data availability}

The datasets used in this study are publicly available. CIFAR-10 and CIFAR-100 are available from the original CIFAR website:
\url{https://www.cs.toronto.edu/~kriz/cifar.html}~\cite{krizhevsky2009learning}. TieredImageNet is available as a publicly used subset of ImageNet introduced by Ren et al.~\cite{ren2018meta}. The tieredImageNet-H hierarchy follows the WordNet-derived hierarchy released by Bertinetto et al.~\cite{ref22}.



\subsection{Computing resources}

All simulations were performed on a computer with an Intel Core i9-13900K CPU and an NVIDIA GeForce RTX 4090 GPU. The simulation code was parallelized using PyTorch’s built-in parallelization to utilize the GPU resources efficiently. In the main CIFAR-100 experiment, one training run of a ResNet-18 model in the 10\% data regime requires approximately 10 minutes.

\newpage
\section{Detailed experimental results}

\subsection{Hierarchical labels align with human grouping task}
\label{app:hier}

To support the cognitive-grounding analysis in Sec.~\ref{sec2.2}, we provide the details of the human grouping task. This task examined whether the semantic hierarchy used in our experiments is aligned with human conceptual judgments. The hierarchy organizes CIFAR-100 fine classes into progressively broader groups, using CIFAR-100-derived superclass labels and five meta-groups formed by merging semantically related superclasses. Participants provided informed consent before the task, and responses were collected anonymously. The study protocol was reviewed and approved by the institutional review board of the authors’ institution.

\begin{figure}[h!]
	\centering
	\includegraphics[width=\textwidth]{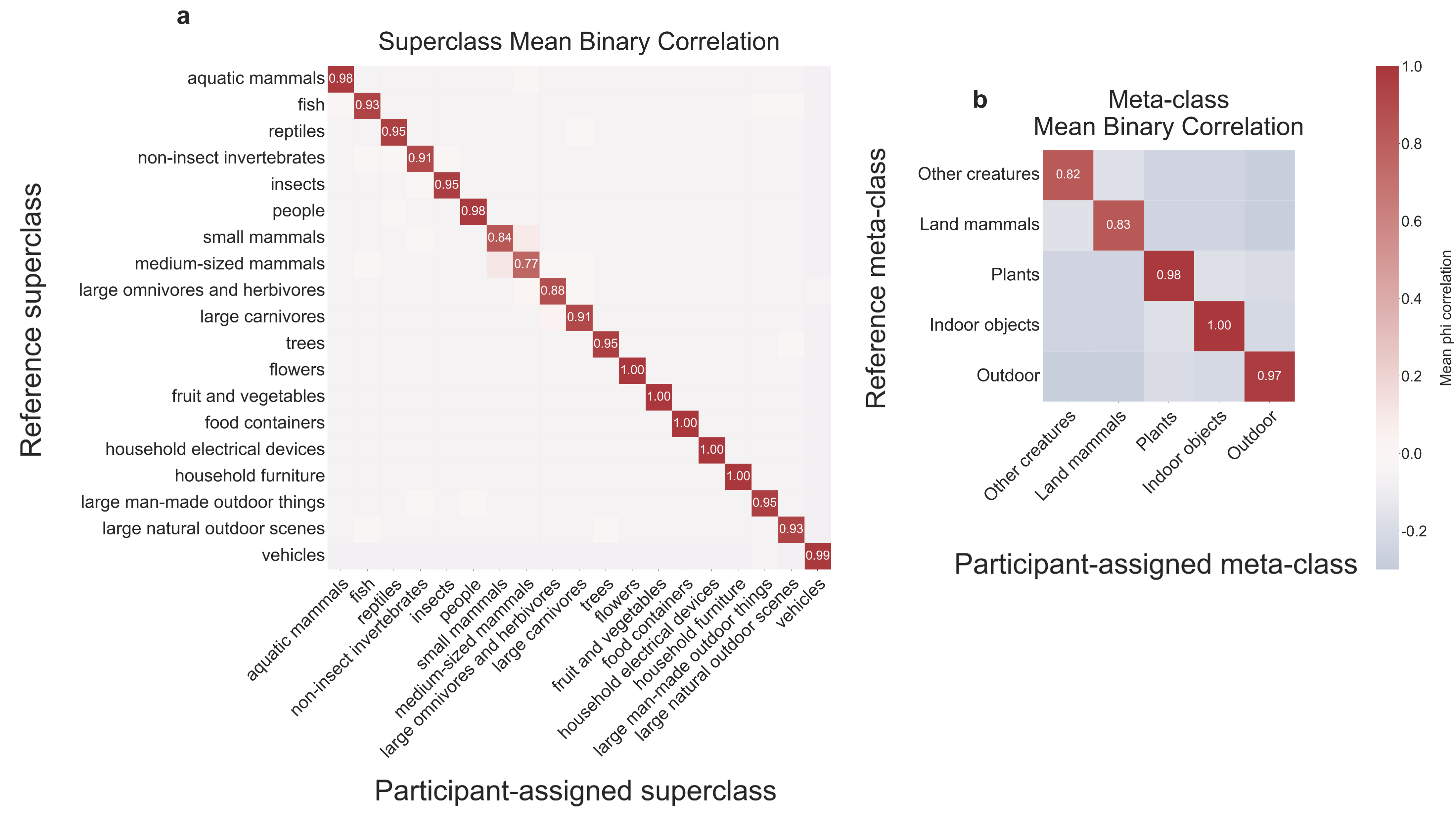}
    \caption{\textbf{Results of human grouping task.}
     (a) Mean binary phi correlations between the CIFAR-100 reference superclass structure and human assignments in the behavioral grouping task. Rows indicate reference superclasses, and columns indicate participant-assigned superclass labels. The strong diagonal pattern shows that participants’ groupings closely matched the predefined CIFAR-100 superclass hierarchy, whereas off-diagonal associations remained near baseline and are therefore left unlabeled. (b) Mean binary phi correlations between the five meta-groups and human assignments in the behavioral grouping task. The strong diagonal pattern shows that participants’ groupings closely matched the meta-level hierarchy constructed by merging semantically related CIFAR-100 superclasses into broader categories.}
	\label{figsb1}
\end{figure}

In the first stage, participants were shown the full set of 100 CIFAR-100 fine-class labels and assigned each label to one of the fixed superclass categories. The category labels were provided in advance, so participants did not generate category names freely. In the second stage, participants assigned the resulting superclass categories to one of five fixed meta-group labels. Thus, the task measured how closely human semantic grouping reproduced the predefined superclass and meta-group structure. Nine valid participants completed the task. Human groupings closely aligned with both levels of the hierarchy: superclass assignments matched the reference structure for 94.78\% $\pm$ 4.55\% of fine classes, far above the task-specific chance expectation of 5.5\% ($\textit{t}$(8) = 58.88, $\textit{p}$ = 7.69 $\times$ 10$^{-12}$), and showed strong pairwise agreement with the predefined superclass structure (mean $\textit{phi}$ = 0.906 $\pm$ 0.084). Meta-group assignments also showed high alignment, with 92.67\% $\pm$ 4.80\% agreement relative to a 21\% chance expectation ($\textit{t}$(8) = 44.83, $\textit{p}$ = 6.77 $\times$ 10$^{-11}$) and strong pairwise correspondence (mean $\textit{phi}$ = 0.815 $\pm$ 0.118). These results indicate that the hierarchy used for this study is strongly aligned with human semantic judgments (Fig.~\ref{figsb1}).

\subsection{Interpretability and attention analysis}
\label{app:attention_analysis}

Here, we provide the selection criteria for the image sets used in the sample-level analysis of Sec.~\ref{sec3.1} and show additional examples corresponding to Fig.~\ref{fig2}d--f. For each image, let $n_{\mathrm{flat}}$ and $n_{\mathrm{SCALA}}$ denote the number of correctly classified models among the 10 runs for the flat-learning model and SCALA, respectively.

"Consensus" samples were defined as images that were reliably classified correctly by both methods:
\begin{equation}
n_{\mathrm{SCALA}} \geq 9
\quad \mathrm{and} \quad
n_{\mathrm{flat}} \geq 9.
\end{equation}

"SCALA-superior" samples were defined as images on which SCALA showed a clear advantage over the flat-learning model:
\begin{equation}
n_{\mathrm{SCALA}} - n_{\mathrm{flat}} \geq 4
\quad \mathrm{and} \quad
n_{\mathrm{SCALA}} \geq 6.
\end{equation}

The symmetric flat-superior criterion was defined as
\begin{equation}
n_{\mathrm{flat}} - n_{\mathrm{SCALA}} \geq 4
\quad \mathrm{and} \quad
n_{\mathrm{flat}} \geq 6.
\end{equation}
However, such cases were rare, with an average of only 0.62 samples per class, and were therefore not analyzed further.

Failure cases were defined as images rarely classified correctly by either method:
\begin{equation}
n_{\mathrm{SCALA}} \leq 1
\quad \mathrm{and} \quad
n_{\mathrm{flat}} \leq 1.
\end{equation}

\begin{figure}[h!]
    \centering
    \includegraphics[width=\textwidth]{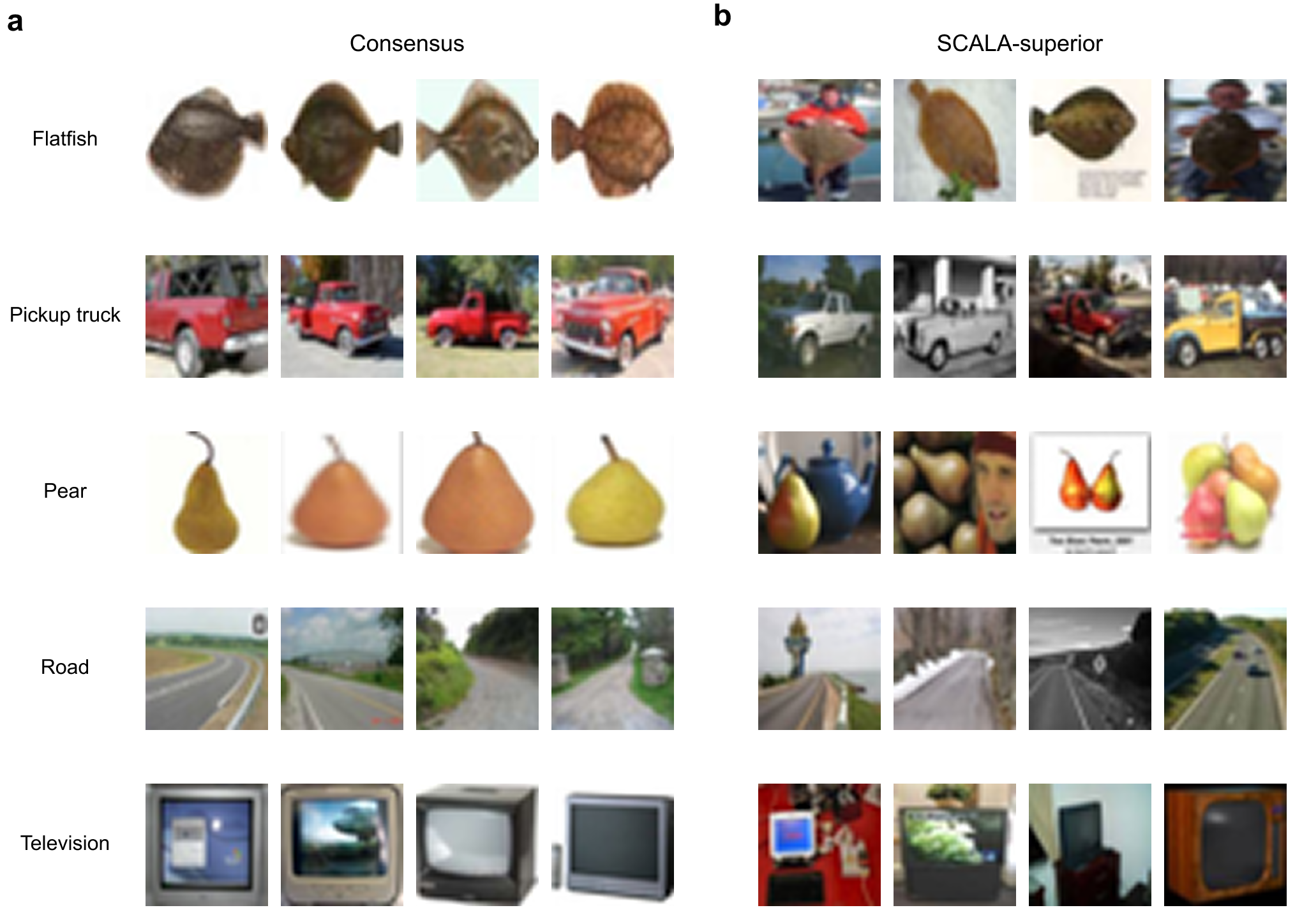}
    \caption{\textbf{Additional consensus and SCALA-superior samples.}
    The examples extend the sample groups shown in Fig.~\ref{fig2}d,e.}
    \label{fig:additional_sample_cases}
\end{figure}

\begin{figure}[h!]
    \centering
    \includegraphics[width=\textwidth]{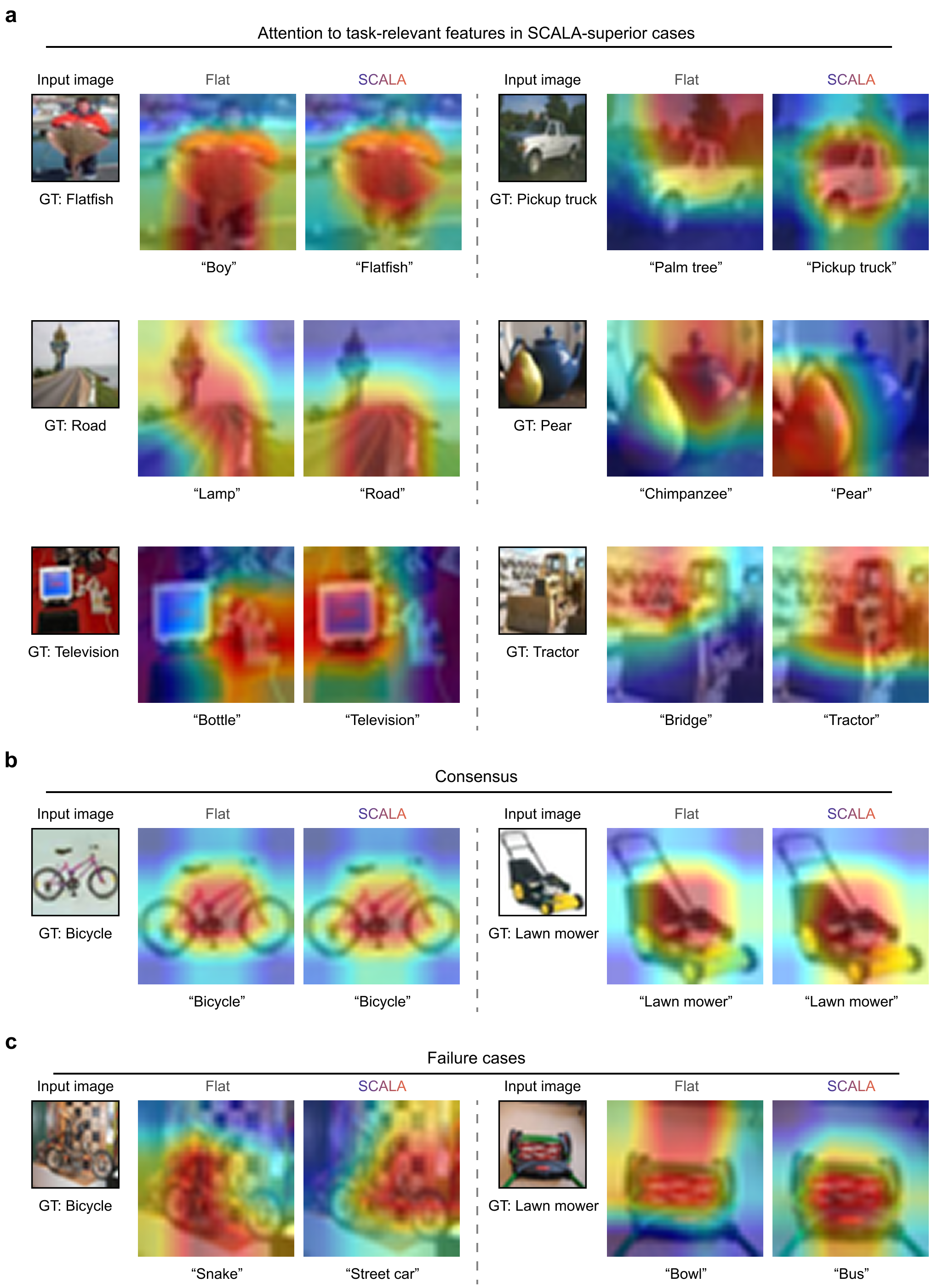}
    \caption{\textbf{Additional Grad-CAM visualizations.}
    Additional examples are shown for (a) SCALA-superior cases, (b) consensus cases, and (c) failure cases.
    Text below each heatmap indicates the model prediction; GT denotes the ground-truth class.}
    \label{fig:additional_gradcam}
\end{figure}

Figure~\ref{fig:additional_sample_cases} shows additional consensus and SCALA-superior image examples, extending the representative samples shown in Fig.~\ref{fig2}d,e. Figure~\ref{fig:additional_gradcam} provides additional Grad-CAM visualizations for SCALA-superior, consensus, and failure cases, extending the attention analysis shown in Fig.~\ref{fig2}f.

\newpage

\subsection{Human perceptual validation of SCALA-superior samples}
\begin{figure}[h!]
	\centering
	\includegraphics[width=\textwidth]{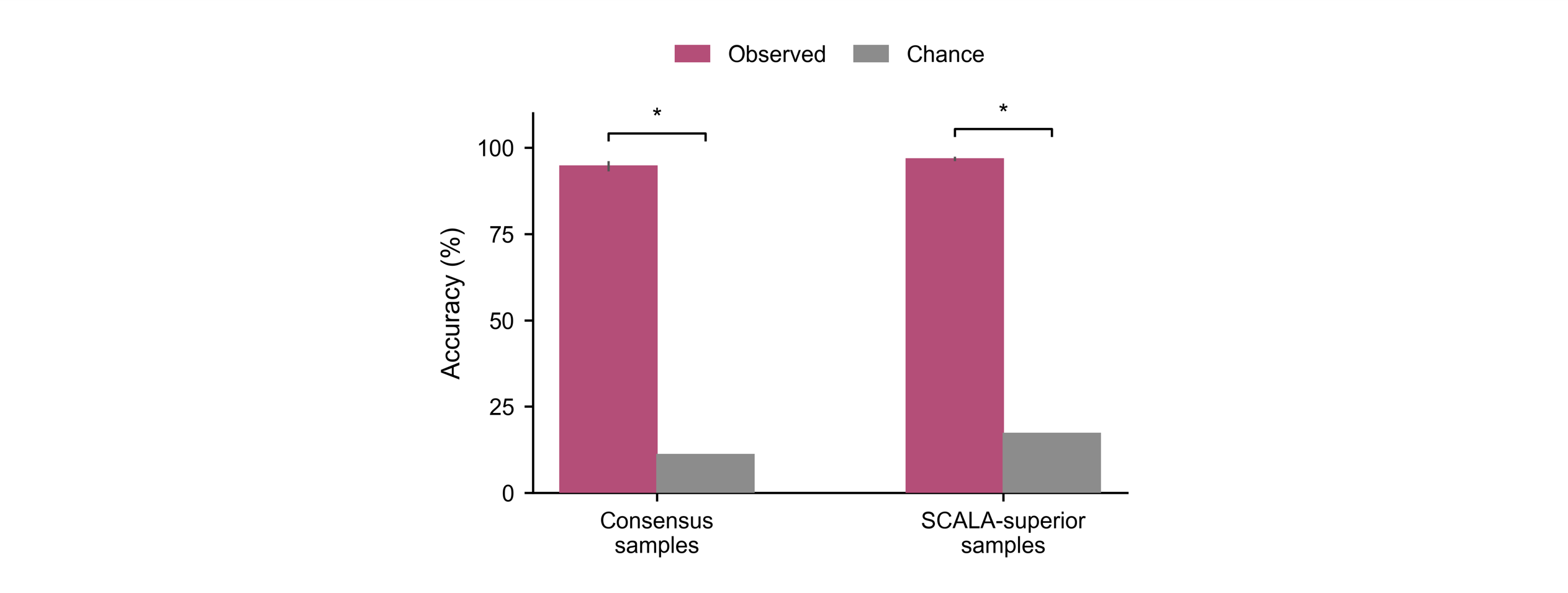}
    \caption{\textbf{Results of human image-classification task.}
     Human observers classified 181 CIFAR-100 images from 15 classes by choosing among candidate fine-class labels that included plausible distractors. Magenta bars indicate observed human fine-class accuracy, and gray bars indicate candidate-set chance. Error bars show SEM across participants for observed accuracy. Asterisks indicate performance significantly above chance (*p < .0001).}
	\label{figsb4}
\end{figure}

To validate the analysis in Sec.~\ref{sec3.1}, we conducted a human image-classification task and report its details here. This task examined whether images on which SCALA outperformed conventional flat learning were also perceptually clear to human observers. Participants voluntarily took part in the task and were compensated after being informed about the nature of the experiment. The stimulus set consisted of 181 CIFAR-100 images from 15 classes, including 119 consensus samples and 62 SCALA-superior samples. Consensus samples were images that both SCALA and flat learning classified reliably, whereas SCALA-superior samples were images for which SCALA outperformed flat learning under the criterion SCALA - flat $\geq$ 4 and SCALA $\geq$ 6. On each trial, participants viewed one image and selected its class from a restricted set of candidate fine-class labels. The candidate set always included the correct target label and plausible distractors, defined as visually or semantically plausible incorrect categories drawn from the model-generated candidate set for that image rather than randomly sampled from all CIFAR-100 classes. Thus, the task tested whether humans could identify the target object even in the presence of model-relevant competing alternatives. Nine participants completed the task. Human performance was near ceiling overall (95.4\%; $\textit{t}$(8) = 78.62, $\textit{p}$ < .0001), and remained high for both consensus samples (94.7\%; $\textit{t}$(8) = 56.17, $\textit{p}$ = 1.12 $\times$ 10$^{-11}$) and SCALA-superior samples (96.8\%; $\textit{t}$(8) = 120.78, $\textit{p}$ = 2.47 $\times$ 10$^{-14}$). Superclass and meta-class accuracy were also near ceiling (99.4\% and 99.6\%, respectively). These results indicate that SCALA-superior samples were not ambiguous or model-specific cases; rather, they were images that humans categorized easily but that flat learning failed to capture as reliably, supporting the interpretation that SCALA better aligns with human perceptual categorization (Fig.~\ref{figsb4}).
\label{Human perceptual validation}

\subsection{Functional upside of broader label grouping}
\label{app:both_wrong_correct}

\subsubsection{Quality of incorrect predictions: Top-5 accuracy and hierarchy consistency}
\label{app:shared_errors_top5}

The failure cases in Fig.~\ref{fig:additional_gradcam}c suggest that even when both models make incorrect fine-class predictions, their errors can differ in semantic severity. We therefore further analyzed samples that both the flat-learning model and SCALA misclassified at the fine-class top-1 level. For these shared-error samples, we evaluated whether the ground-truth fine label remained within the top-5 predictions and whether the top-1 prediction preserved the correct superclass or meta-class.

SCALA showed higher top-5 accuracy than the flat-learning model, indicating that the correct fine class more often remained among its high-confidence alternatives even when the top-1 prediction was incorrect. SCALA also achieved higher superclass and meta-class consistency, suggesting that its errors were more likely to remain within the correct higher-level semantic category rather than shifting to unrelated classes.

\begin{table}[h!]
  \caption{
  Top-5 accuracy and hierarchy-level consistency on samples that both models misclassified. Each value (\%) is reported as mean $\pm$ standard deviation across ten trials.}
  \label{tab:shared_errors_top5}
  \centering
  \setlength{\tabcolsep}{6pt}
  \begin{tabular}{lccc}
    \toprule
    Metric
    & Flat [\%]
    & SCALA [\%]
    & $\Delta$ \\
    \midrule
    Top-5 accuracy
    & $34.8 \pm 1.7$
    & $42.4 \pm 2.0$
    & $\blacktriangle\, 7.5 \pm 3.1$ \\
    Superclass consistency
    & $21.4 \pm 0.8$
    & $24.8 \pm 1.6$
    & $\blacktriangle\, 3.4 \pm 1.4$ \\
    Meta-class consistency
    & $51.6 \pm 1.3$
    & $56.0 \pm 1.9$
    & $\blacktriangle\, 4.4 \pm 2.1$ \\
    \bottomrule
  \end{tabular}
\end{table}

\subsubsection{Quality of correct predictions: representation and confidence}
\label{app:confidence_analysis}

\begin{figure}[H]
	\centering
    \includegraphics[width=\textwidth]{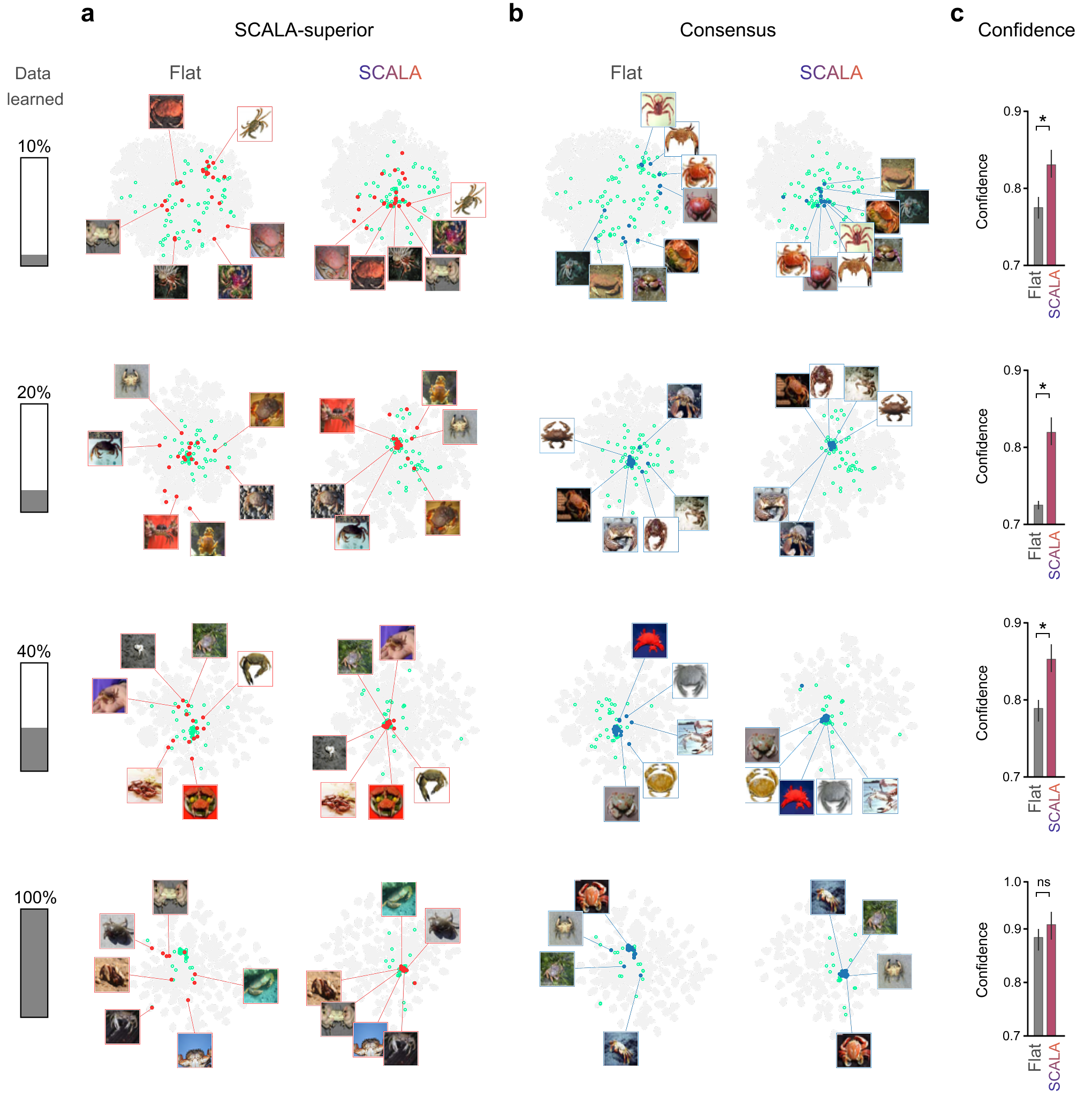}
    \caption{\textbf{Feature-space organization and confidence in selected sample groups.}
    (a,b) Final representation-space visualizations for the crab class across different training-data regimes, showing samples correctly classified by SCALA but misclassified by the flat-learning model (red points; a) and samples correctly classified by both models (blue points; b).
    Cyan points indicate crab-class samples and gray points indicate other test samples. Representative images are overlaid on selected points.
    (c) Prediction confidence on consensus samples across all classes.}
    \label{fig:confidence_analysis}
\end{figure}

We next examined how the sample groups analyzed in Sec.~\ref{sec3.1} were organized in the final representation space by visualizing penultimate-layer features with t-SNE. Figure~\ref{fig:confidence_analysis}a shows samples that were correctly classified by SCALA but misclassified by the flat-learning model. Consistent with the SCALA-superior examples in Fig.~\ref{fig2}e, these samples often contained more complex visual evidence, such as cluttered backgrounds, non-canonical object appearances, or competing objects. In the flat-learning model, these samples were more dispersed in the feature space, whereas in SCALA they formed a more coherent local structure around the corresponding class region.

We then analyzed consensus samples that were correctly classified by both models (Fig.~\ref{fig:confidence_analysis}b). These images were generally more canonical and visually less ambiguous, consistent with the examples in Fig.~\ref{fig2}d. However, even when both models predicted the correct label, the flat-learning model produced a more fragmented organization, with several samples remaining as outliers from the class cluster. In contrast, SCALA formed a more compact cluster for the same consensus samples. This provides a sample-level counterpart to the representation-geometry results in Sec.~\ref{sec3.2}, where SCALA produced more compact and semantically organized representations.

Finally, we compared prediction confidence on consensus samples, where both models produced correct predictions (Fig.~\ref{fig:confidence_analysis}c). Confidence was measured as the softmax probability assigned to the ground-truth fine class; because all samples in this analysis were correctly classified by both models, this is also the probability assigned to the predicted class. In this setting, higher confidence indicates that the correct prediction was made with stronger probability mass. SCALA showed significantly higher confidence than the flat-learning model in data-limited regimes (Flat vs. SCALA, $n_{\text{net}}=10$, Wilcoxon rank-sum test, $\textit{p}<0.05$), while the difference was not significant in the full-data setting. This suggests that hierarchical scaffolding induces greater predictive consistency in data-limited regimes, even for samples accurately classified by both models.


\subsection{Learning dynamics and performance across dataset sizes}

\begin{figure}[H]
	\centering
	\includegraphics[width=\textwidth]{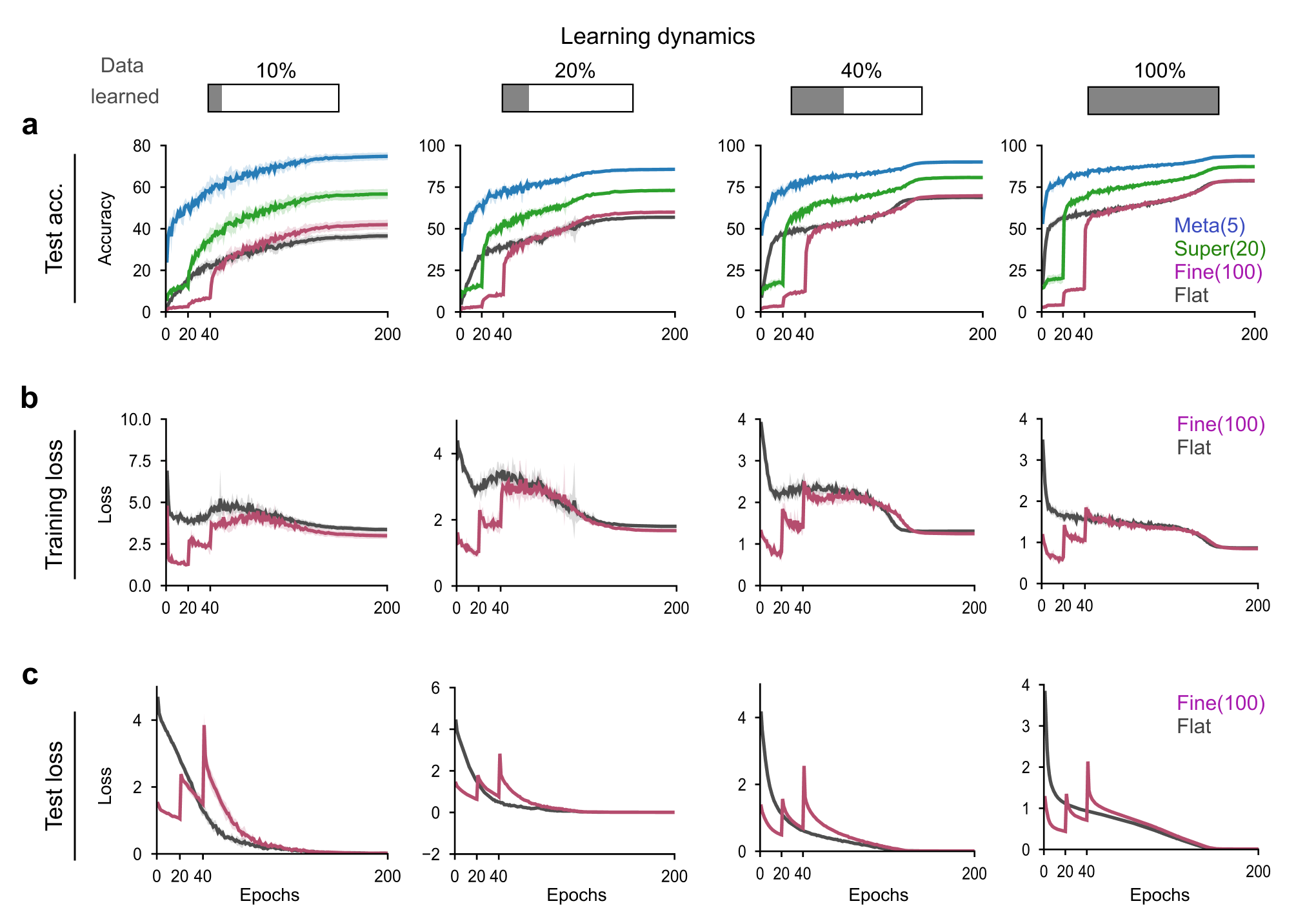}
    \caption{\textbf{Learning dynamics across training set sizes.}
    Results are shown for training set sizes of 10\%, 20\%, 40\%, and 100\% of the original CIFAR-100 training set.
    (a) Test accuracy over training. For SCALA, accuracy is reported at the meta (blue), superclass (green), and fine (magenta) levels; the flat-learning model fine-level accuracy is shown in black.
    (b) Fine-level training loss over epochs for SCALA (magenta) and the flat-learning model (black).
    (c) Fine-level test loss over epochs for SCALA (magenta) and the flat-learning model (black).}
	\label{fig:learning_dynamics_dataset_sizes}
\end{figure}

We further examined learning dynamics across training set sizes (Fig.~\ref{fig:learning_dynamics_dataset_sizes}). SCALA shows stage-wise behavior: meta-level accuracy improves during the initial coarse stage, superclass accuracy increases after the transition to superclass-level supervision, and fine-class accuracy rises rapidly once fine-level supervision begins. Although the fine-level loss exhibits transient changes around stage transitions, SCALA converges to competitive or improved fine-level performance relative to the flat-learning model. Final performance across data regimes and evaluation levels confirmed that SCALA outperformed the flat-learning model at the meta-, super-, and fine-class levels, with larger gains as the amount of training data decreased (Table~\ref{tab:performance_dataset_sizes}).

\begin{table}[t]
    \centering
    \caption{Performance under different training set sizes for the flat-learning model and SCALA, evaluated at the meta-, super-, and fine-class levels. Each value (\%) shows mean $\pm$ standard deviation across ten trials.}
    \label{tab:performance_dataset_sizes}
    \setlength{\tabcolsep}{3.5pt}
    \resizebox{\textwidth}{!}{%
    \begin{tabular}{lccc ccc ccc}
    \toprule
    & \multicolumn{3}{c}{Meta Accuracy ($N_{\mathrm{class}}=5$)}
    & \multicolumn{3}{c}{Super Accuracy ($N_{\mathrm{class}}=20$)}
    & \multicolumn{3}{c}{Fine Accuracy ($N_{\mathrm{class}}=100$)} \\
    \cmidrule(lr){2-4} \cmidrule(lr){5-7} \cmidrule(lr){8-10}
    Data
    & Flat [\%] & SCALA [\%] & $\Delta$
    & Flat [\%] & SCALA [\%] & $\Delta$
    & Flat [\%] & SCALA [\%] & $\Delta$ \\
    \midrule
    10\%
    & 69.7$\,\pm\,1.5$ & 74.8$\,\pm\,2.0$ & $\blacktriangle\ \mathbf{5.1\pm2.2}$
    & 50.6$\,\pm\,1.7$ & 56.8$\,\pm\,2.6$ & $\blacktriangle\ \mathbf{6.2\pm2.7}$
    & 36.7$\,\pm\,1.8$ & 42.2$\,\pm\,2.5$ & $\blacktriangle\ \mathbf{5.5\pm2.8}$ \\
    
    20\%
    & 83.7$\,\pm\,0.4$ & 85.6$\,\pm\,0.3$ & $\blacktriangle\ \mathbf{1.9\pm0.5}$
    & 69.9$\,\pm\,0.6$ & 73.1$\,\pm\,0.4$ & $\blacktriangle\ \mathbf{3.2\pm0.7}$
    & 57.1$\,\pm\,0.7$ & 60.2$\,\pm\,0.3$ & $\blacktriangle\ \mathbf{3.0\pm0.9}$ \\

    40\%
    & 89.4$\,\pm\,0.3$ & 90.1$\,\pm\,0.2$ & $\blacktriangle\ \mathbf{0.7\pm0.3}$
    & 79.8$\,\pm\,0.4$ & 80.8$\,\pm\,0.4$ & $\blacktriangle\ \mathbf{1.1\pm0.5}$
    & 69.0$\,\pm\,0.4$ & 70.0$\,\pm\,0.3$ & $\blacktriangle\ \mathbf{1.0\pm0.5}$ \\

    100\%
    & 93.5$\,\pm\,0.2$ & 93.5$\,\pm\,0.1$ & $\blacktriangle\ \mathbf{0.1\pm0.2}$
    & 87.2$\,\pm\,0.2$ & 87.4$\,\pm\,0.2$ & $\blacktriangle\ \mathbf{0.2\pm0.3}$
    & 78.9$\,\pm\,0.3$ & 79.1$\,\pm\,0.1$ & $\blacktriangle\ \mathbf{0.1\pm0.3}$ \\
    \bottomrule
    \end{tabular}%
    }
\end{table}

\subsection{Representation geometry analysis}



\subsubsection{Metrics for final representation geometry}
\label{app:evaluation_metrics}

This section defines the clustering metrics used to quantify final representation geometry in Sec.~\ref{sec3.2}.
We evaluated representation geometry at the meta-, super-, and fine-class levels. For each semantic level $m$, clusters were defined by the corresponding labels. Let $h_i$ denote the penultimate-layer feature of sample $i$, and let $\mathcal{C}^{(m)}_k$ denote the set of samples assigned to cluster $k$ at level $m$.

\paragraph{Intra-class distance.}
We measured intra-class compactness using the within-cluster distance term from the silhouette formulation~\cite{rousseeuw1987silhouettes}. For each sample $i$, we computed the average distance to all other samples in the same cluster:
\begin{equation}
a_i^{(m)}
=
\frac{1}{|\mathcal{C}^{(m)}_{y_i}|-1}
\sum_{\substack{j \in \mathcal{C}^{(m)}_{y_i} \\ j \neq i}}
\|h_i - h_j\|_2 .
\end{equation}
The intra-class distance at level $m$ was then computed as
\begin{equation}
D_{\mathrm{intra}}^{(m)}
=
\frac{1}{n}
\sum_{i=1}^{n}
a_i^{(m)} .
\end{equation}

\paragraph{Calinski--Harabasz Index (CHI).}
We used the Calinski--Harabasz index (CHI) to quantify global cluster separability~\cite{calinski1974dendrite}. Let $\mu$ denote the global feature centroid, $\mu_k^{(m)}$ denote the centroid of cluster $k$ at level $m$, $n_k$ denote the number of samples in cluster $k$, and $K_m$ denote the number of clusters. CHI is defined as
\begin{equation}
\mathrm{CHI}^{(m)}
=
\frac{
\sum_{k=1}^{K_m} n_k \|\mu_k^{(m)} - \mu\|_2^2 / (K_m - 1)
}{
\sum_{k=1}^{K_m} \sum_{i \in \mathcal{C}^{(m)}_k}
\|h_i - \mu_k^{(m)}\|_2^2 / (n - K_m)
}.
\end{equation}

\paragraph{Davies--Bouldin Index (DBI).}
We used the Davies--Bouldin index (DBI) to measure local cluster overlap~\cite{davies1979cluster}. For each cluster $k$, we first computed the average within-cluster dispersion
\begin{equation}
S_k^{(m)}
=
\frac{1}{n_k}
\sum_{i \in \mathcal{C}^{(m)}_k}
\|h_i - \mu_k^{(m)}\|_2 .
\end{equation}
The pairwise similarity between clusters $k$ and $l$ was then defined as
\begin{equation}
R_{kl}^{(m)}
=
\frac{
S_k^{(m)} + S_l^{(m)}
}{
\|\mu_k^{(m)} - \mu_l^{(m)}\|_2
}.
\end{equation}
DBI is computed as
\begin{equation}
\mathrm{DBI}^{(m)}
=
\frac{1}{K_m}
\sum_{k=1}^{K_m}
\max_{l \neq k} R_{kl}^{(m)} .
\end{equation}

\subsubsection{Progressive representation formation during training}

We provide an additional example of progressive representation formation in the 10\% data regime, using a different superclass from the main analysis. In addition to the t-SNE trajectory, we report CKA similarity to the full-data reference across all four residual stages of ResNet-18. 

\begin{figure}[h!]
	\centering
	\includegraphics[width=\textwidth]{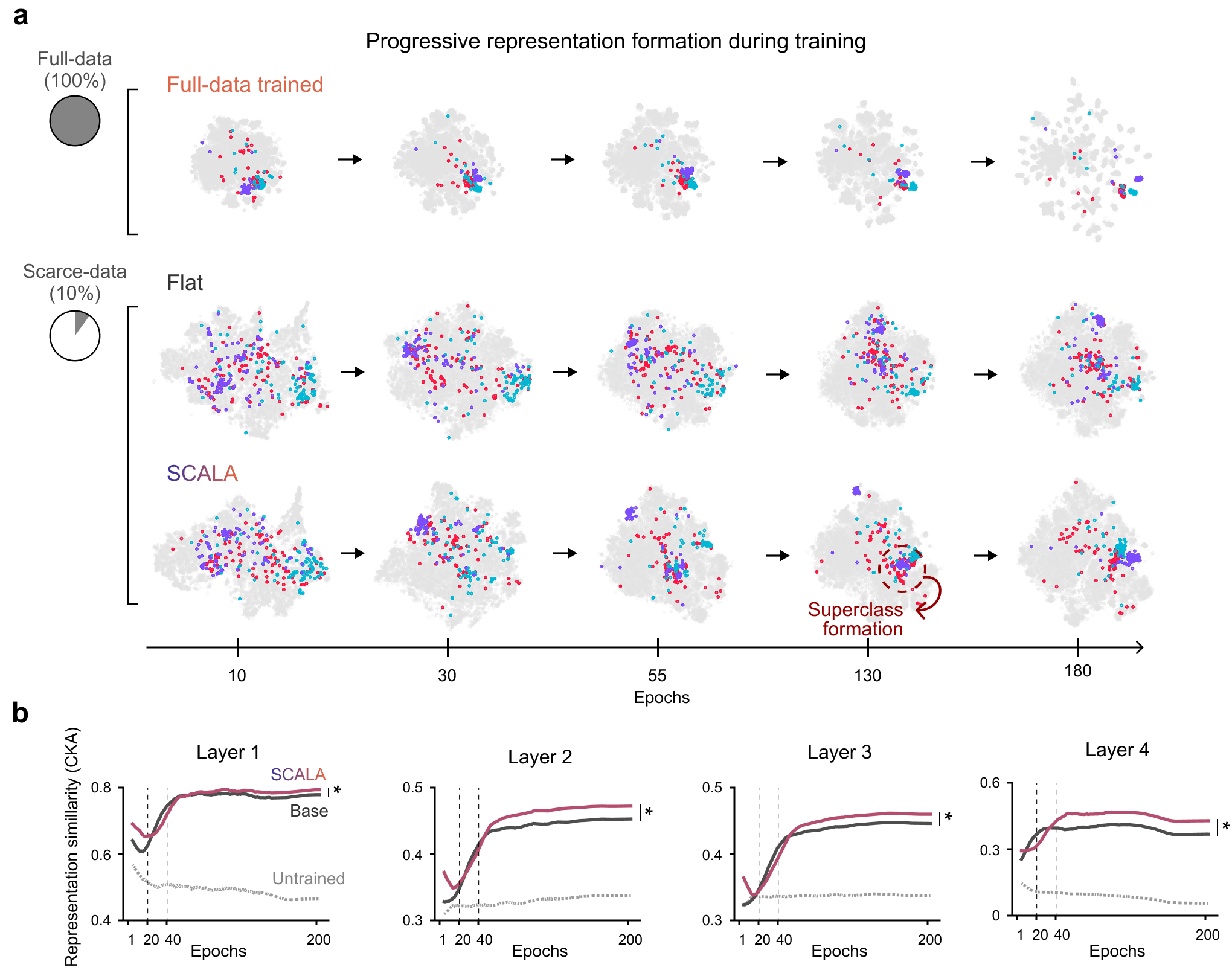}
    \caption{\textbf{Progressive representation formation.}
    (a) Evolution of representation geometry visualized with t-SNE in a 10\% data regime, shown for an additional superclass example. The flat-learning model and SCALA are compared across epochs 10, 30, 55, 130, and 180, with the full-data model provided as a reference. Fine classes within the \textit{household\_furniture} superclass are highlighted: \textit{chair} (cyan), \textit{bed} (red), and \textit{wardrobe} (purple). Gray points indicate all other classes.
    (b) Representational similarity to the full-data reference across all four residual stages of ResNet-18. The solid lines show the similarity trajectories for SCALA and the flat-learning model, while the gray dotted line indicates an untrained network, which serves as a lower-bound reference. The dashed vertical lines indicate the stage transitions at epochs 20 and 40. All layer-wise comparisons between SCALA and the flat-learning model were statistically significant (Flat vs. SCALA, $n_{\mathrm{net}}=10$, Wilcoxon rank-sum test, $p<0.05$).}
	\label{figs1}
\end{figure}


\subsubsection{Metrics for progressive representation formation}
\label{app:cka}

For the progressive representation analysis in Sec.~\ref{sec3.3}, we quantified two aspects of representation formation: similarity to a full-data reference using CKA, and fine-class compactness using cosine intra-class distance. Both metrics were computed on the same evaluation samples across training epochs.

We used cosine distance for this epoch-wise analysis, as the overall scale of penultimate-layer features can change substantially over training. By reducing the influence of feature-magnitude variation, this metric emphasizes the relative organization of class representations.

\paragraph{Centered Kernel Alignment (CKA).} We used the linear centered kernel alignment (CKA) formulation in Kornblith et al.~\cite{ref27}. For each layer $\ell$, activations were extracted from the same set of input images for the scarce-data model and the full-data reference model. Let $X_{\ell} \in \mathbb{R}^{n \times d_{\ell}}$ and $Y_{\ell} \in \mathbb{R}^{n \times d_{\ell}}$ denote the corresponding centered activation matrices. Linear CKA was computed as 
\begin{equation}
\mathrm{CKA}(X_{\ell}, Y_{\ell})
=
\frac{
\left\| X_{\ell}^{\top} Y_{\ell} \right\|_{F}^{2}
}{
\left\| X_{\ell}^{\top} X_{\ell} \right\|_{F}
\left\| Y_{\ell}^{\top} Y_{\ell} \right\|_{F}
}.
\end{equation}
Here, $\|\cdot\|_{F}$ denotes the Frobenius norm. Higher CKA values indicate greater representational similarity between the two models at the corresponding layer.

\paragraph{Cosine intra-class distance.}
To quantify progressive compactness during training, we also measured intra-class distance using cosine distance between penultimate-layer features. For samples $i$ and $j$, cosine distance was defined as
\begin{equation}
d_{\cos}(h_i,h_j)
=
1 -
\frac{h_i^\top h_j}{\|h_i\|_2 \|h_j\|_2}.
\end{equation}
For each sample $i$, we computed the average cosine distance to all other samples in the same fine class:
\begin{equation}
a_{i,\cos}
=
\frac{1}{|\mathcal{C}_{y_i}|-1}
\sum_{\substack{j \in \mathcal{C}_{y_i} \\ j \neq i}}
d_{\cos}(h_i,h_j),
\end{equation}
where $\mathcal{C}_{y_i}$ denotes the set of samples with the same fine-class label as sample $i$. The cosine intra-class distance was then averaged over all samples:
\begin{equation}
D_{\mathrm{cos}}
=
\frac{1}{n}
\sum_{i=1}^{n}
a_{i,\cos}.
\end{equation}

\subsection{Hierarchical scaffolding supports generalization and adaptation to unseen classes}
\label{app:unseen_class_results}

We report the full set of quantitative results for the unseen-class experiment summarized in Fig.~\ref{fig5}. Table~\ref{tab:unseen_zero_shot} expands the zero-shot higher-level generalization result in Fig.~\ref{fig5}c across all training set sizes. Table~\ref{tab:unseen_after_adaptation} expands the novel-class adaptation result in Fig.~\ref{fig5}d by reporting meta-, super-, and fine-level performance on the reintroduced 20 classes. Table~\ref{tab:seen80_before_adaptation} reports the performance of the 80-class checkpoint at the end of the initial learning phase, corresponding to the pre-adaptation state in Fig.~\ref{fig5}b. Tables~\ref{tab:seen80_after_adaptation} and~\ref{tab:all100_after_adaptation} report performance after adaptation on the previously seen 80 classes and on the full 100-class set, respectively, complementing the adaptation phase shown in Fig.~\ref{fig5}b,d.


\begin{table}[h!]
\centering
\caption{Zero-shot generalization performance on unseen classes at the meta and super levels under different training set sizes. Each value (\%) shows mean $\pm$ standard deviation across ten trials.}
\label{tab:unseen_zero_shot}
\footnotesize
\setlength{\tabcolsep}{4pt}
\renewcommand{\arraystretch}{0.95}
\begin{tabular}{lccc ccc}
\toprule
& \multicolumn{3}{c}{Meta Accuracy ($N_{\mathrm{class}}=5$)}
& \multicolumn{3}{c}{Super Accuracy ($N_{\mathrm{class}}=20$)} \\
\cmidrule(lr){2-4} \cmidrule(lr){5-7}
data & Flat [\%] & SCALA [\%] & $\Delta$
     & Flat [\%] & SCALA [\%] & $\Delta$ \\
\midrule
10\%  & 54.7$\pm$1.8 & 58.2$\pm$1.5 & $\blacktriangle\ 3.6\pm2.1$
  & 26.9$\pm$1.6 & 30.4$\pm$1.1 & $\blacktriangle\ 3.6\pm1.8$ \\
20\%  & 65.8$\pm$0.8 & 68.0$\pm$1.9 & $\blacktriangle\ 2.2\pm1.9$
  & 37.3$\pm$1.0 & 39.5$\pm$1.6 & $\blacktriangle\ 2.2\pm1.3$ \\
40\%  & 71.1$\pm$0.8 & 72.4$\pm$1.1 & $\blacktriangle\ 1.3\pm0.7$
      & 43.2$\pm$0.5 & 43.9$\pm$0.8 & $\blacktriangle\ 0.7\pm0.9$ \\
100\% & 74.5$\pm$0.8 & 75.5$\pm$0.9 & $\blacktriangle\ 1.0\pm1.5$
  & 46.9$\pm$0.8 & 48.1$\pm$0.8 & $\blacktriangle\ 1.1\pm1.3$ \\
\bottomrule
\end{tabular}
\end{table}


\begin{table}[h!]
  \caption{Performance on unseen classes after reintroducing the held-out categories and performing additional training, evaluated at the meta, super, and fine levels under different training set sizes. Each value (\%) shows mean $\pm$ standard deviation across ten trials.}
  \label{tab:unseen_after_adaptation}
  \centering

  \footnotesize
  \setlength{\tabcolsep}{3pt}
  \resizebox{\columnwidth}{!}{%
  \begin{tabular}{lccc ccc ccc}
    \toprule
    & \multicolumn{3}{c}{Meta Accuracy ($N_{\mathrm{class}}=5$)}
    & \multicolumn{3}{c}{Super Accuracy ($N_{\mathrm{class}}=20$)}
    & \multicolumn{3}{c}{Fine Accuracy ($N_{\mathrm{class}}=100$)} \\
    \cmidrule(lr){2-4} \cmidrule(lr){5-7} \cmidrule(lr){8-10}
    data
    & Flat [\%] & SCALA [\%] & $\Delta$
    & Flat [\%] & SCALA [\%] & $\Delta$
    & Flat [\%] & SCALA [\%] & $\Delta$ \\
    \midrule
    10\%
    & 70.3$\pm$1.6 & 74.4$\pm$1.8 & $\blacktriangle\ 4.0\pm2.3$
    & 51.4$\pm$1.9 & 57.1$\pm$1.9 & $\blacktriangle\ 5.7\pm2.4$
    & 38.5$\pm$1.5 & 47.6$\pm$1.6 & $\blacktriangle\ 9.1\pm2.2$ \\
    20\%
    & 82.6$\pm$0.8 & 84.1$\pm$0.6 & $\blacktriangle\ 1.5\pm0.6$
    & 69.0$\pm$0.8 & 71.5$\pm$0.7 & $\blacktriangle\ 2.5\pm0.8$
    & 57.1$\pm$0.8 & 62.5$\pm$0.9 & $\blacktriangle\ 5.4\pm0.9$ \\
    40\%
    & 87.8$\pm$0.6 & 88.8$\pm$0.6 & $\blacktriangle\ 1.1\pm0.8$
    & 77.7$\pm$0.6 & 79.3$\pm$0.6 & $\blacktriangle\ 1.6\pm0.8$
    & 67.7$\pm$0.6 & 69.8$\pm$0.8 & $\blacktriangle\ 2.2\pm1.0$ \\
    100\%
    & 91.4$\pm$0.4 & 92.0$\pm$0.5 & $\blacktriangle\ 0.6\pm0.7$
    & 84.4$\pm$0.5 & 85.1$\pm$0.5 & $\blacktriangle\ 0.7\pm0.8$
    & 76.4$\pm$0.4 & 77.0$\pm$0.6 & $\blacktriangle\ 0.5\pm0.7$ \\    
    \bottomrule
  \end{tabular}
  }%
\end{table}


\begin{table}[h!]
  \caption{Performance after training on 80 classes, evaluated at the meta-, super-, and fine-class levels under different training set sizes. Each value (\%) shows mean $\pm$ standard deviation across ten trials.}
  \label{tab:seen80_before_adaptation}
  \centering
  \setlength{\tabcolsep}{3.5pt}
  \resizebox{\columnwidth}{!}{%
  \begin{tabular}{lccc ccc ccc}
    \toprule
    & \multicolumn{3}{c}{Meta-class Accuracy ($N_{\mathrm{class}}=5$)}
    & \multicolumn{3}{c}{Superclass Accuracy ($N_{\mathrm{class}}=20$)}
    & \multicolumn{3}{c}{Fine-class Accuracy ($N_{\mathrm{class}}=100$)} \\
    \cmidrule(lr){2-4} \cmidrule(lr){5-7} \cmidrule(lr){8-10}
    data
    & Flat [\%] & SCALA [\%] & $\Delta$
    & Flat [\%] & SCALA [\%] & $\Delta$
    & Flat [\%] & SCALA [\%] & $\Delta$ \\
    \midrule
    10\%
    & 73.5$\pm$1.3 & 76.7$\pm$2.0 & $\blacktriangle\ 3.2\pm2.0$
    & 56.4$\pm$1.6 & 60.5$\pm$2.3 & $\blacktriangle\ 4.1\pm2.3$
    & 44.5$\pm$1.7 & 48.3$\pm$2.4 & $\blacktriangle\ 3.8\pm2.4$ \\
    20\%
    & 85.9$\pm$0.4 & 87.3$\pm$0.4 & $\blacktriangle\ 1.5\pm0.4$
    & 73.9$\pm$0.6 & 76.0$\pm$0.6 & $\blacktriangle\ 2.1\pm0.6$
    & 63.9$\pm$0.8 & 65.8$\pm$0.8 & $\blacktriangle\ 1.9\pm0.6$ \\
    40\%
    & 91.2$\pm$0.3 & 91.5$\pm$0.3 & $\blacktriangle\ 0.3\pm0.4$
    & 82.8$\pm$0.4 & 83.5$\pm$0.4 & $\blacktriangle\ 0.7\pm0.5$
    & 74.5$\pm$0.4 & 75.0$\pm$0.3 & $\blacktriangle\ 0.5\pm0.4$ \\
    100\%
    & 93.8$\pm$0.2 & 94.0$\pm$0.1 & $\blacktriangle\ 0.2\pm0.2$
    & 87.8$\pm$0.2 & 88.2$\pm$0.3 & $\blacktriangle\ 0.4\pm0.3$
    & 81.1$\pm$0.2 & 81.3$\pm$0.2 & $\blacktriangle\ 0.2\pm0.2$ \\
    \bottomrule
  \end{tabular}%
  }
\end{table}

\label{app:seen_unseen}

\begin{table}[h!]
  \caption{Performance on previously seen classes after reintroducing the held-out categories and performing additional training, evaluated at the meta-, super-, and fine-class levels under different training set sizes. Each value (\%) shows mean $\pm$ standard deviation across ten trials.}
  \label{tab:seen80_after_adaptation}
  \centering
  \setlength{\tabcolsep}{3.5pt}
  \resizebox{\columnwidth}{!}{%
  \begin{tabular}{lccc ccc ccc}
    \toprule
    & \multicolumn{3}{c}{Meta-class Accuracy ($N_{\mathrm{class}}=5$)}
    & \multicolumn{3}{c}{Superclass Accuracy ($N_{\mathrm{class}}=20$)}
    & \multicolumn{3}{c}{Fine-class Accuracy ($N_{\mathrm{class}}=100$)} \\
    \cmidrule(lr){2-4} \cmidrule(lr){5-7} \cmidrule(lr){8-10}
    data
    & Flat [\%] & SCALA [\%] & $\Delta$
    & Flat [\%] & SCALA [\%] & $\Delta$
    & Flat [\%] & SCALA [\%] & $\Delta$ \\
    \midrule
    10\%
    & 75.3$\pm$1.5 & 78.5$\pm$1.5 & $\blacktriangle\ 3.1\pm1.7$
    & 57.6$\pm$2.0 & 61.5$\pm$2.0 & $\blacktriangle\ 3.9\pm2.1$
    & 42.6$\pm$1.8 & 43.0$\pm$2.1 & $\blacktriangle\ 0.3\pm2.3$ \\
    20\%
    & 85.7$\pm$0.3 & 86.4$\pm$0.5 & $\blacktriangle\ 0.7\pm0.5$
    & 73.4$\pm$0.5 & 74.2$\pm$0.7 & $\blacktriangle\ 0.8\pm0.7$
    & 60.5$\pm$0.7 & 59.4$\pm$0.8 & $\blacktriangledown\ 1.2\pm0.7$ \\
    40\%
    & 90.2$\pm$0.3 & 90.2$\pm$0.2 & $\blacktriangle\ 0.1\pm0.4$
    & 80.8$\pm$0.4 & 81.1$\pm$0.2 & $\blacktriangle\ 0.3\pm0.4$
    & 70.2$\pm$0.5 & 69.3$\pm$0.2 & $\blacktriangledown\ 0.9\pm0.5$ \\
    100\%
    & 92.8$\pm$0.3 & 92.9$\pm$0.2 & $\blacktriangle\ 0.1\pm0.4$
    & 86.1$\pm$0.3 & 86.1$\pm$0.2 & $\blacktriangle\ 0.0\pm0.4$
    & 77.1$\pm$0.3 & 76.8$\pm$0.3 & $\blacktriangledown\ 0.3\pm0.4$ \\
    \bottomrule
  \end{tabular}%
  }
\end{table}

\begin{table}[h!]
  \caption{Performance on all classes after reintroducing the held-out categories and performing additional training, evaluated at the meta-, super-, and fine-class levels under different training set sizes. Each value (\%) shows mean $\pm$ standard deviation across ten trials.}
  \label{tab:all100_after_adaptation}
  \centering
  \setlength{\tabcolsep}{3.5pt}
  \resizebox{\columnwidth}{!}{%
  \begin{tabular}{lccc ccc ccc}
    \toprule
    & \multicolumn{3}{c}{Meta-class Accuracy ($N_{\mathrm{class}}=5$)}
    & \multicolumn{3}{c}{Superclass Accuracy ($N_{\mathrm{class}}=20$)}
    & \multicolumn{3}{c}{Fine-class Accuracy ($N_{\mathrm{class}}=100$)} \\
    \cmidrule(lr){2-4} \cmidrule(lr){5-7} \cmidrule(lr){8-10}
    data
    & Flat [\%] & SCALA [\%] & $\Delta$
    & Flat [\%] & SCALA [\%] & $\Delta$
    & Flat [\%] & SCALA [\%] & $\Delta$ \\
    \midrule
    10\%
    & 74.3$\pm$1.5 & 77.7$\pm$1.5 & $\blacktriangle\ 3.3\pm1.8$
    & 56.3$\pm$1.9 & 60.6$\pm$1.9 & $\blacktriangle\ 4.3\pm2.2$
    & 41.8$\pm$1.7 & 43.9$\pm$2.0 & $\blacktriangle\ 2.1\pm2.2$ \\
    20\%
    & 85.1$\pm$0.3 & 85.9$\pm$0.4 & $\blacktriangle\ 0.9\pm0.4$
    & 72.5$\pm$0.4 & 73.6$\pm$0.6 & $\blacktriangle\ 1.2\pm0.6$
    & 59.8$\pm$0.6 & 60.0$\pm$0.6 & $\blacktriangle\ 0.2\pm0.6$ \\
    40\%
    & 89.7$\pm$0.3 & 89.9$\pm$0.2 & $\blacktriangle\ 0.3\pm0.3$
    & 80.2$\pm$0.3 & 80.7$\pm$0.3 & $\blacktriangle\ 0.5\pm0.3$
    & 69.7$\pm$0.4 & 69.4$\pm$0.2 & $\blacktriangledown\ 0.3\pm0.3$ \\
    100\%
    & 92.6$\pm$0.3 & 92.7$\pm$0.2 & $\blacktriangle\ 0.2\pm0.3$
    & 85.8$\pm$0.3 & 85.9$\pm$0.2 & $\blacktriangle\ 0.2\pm0.3$
    & 77.0$\pm$0.3 & 76.9$\pm$0.3 & $\blacktriangledown\ 0.1\pm0.3$ \\
    \bottomrule
  \end{tabular}%
  }
\end{table}

\subsection{Analysis across training configurations}

\subsubsection{Number of hierarchy stages}

\begin{table}[h!]
\caption{Performance under different training set sizes and hierarchy-stage schedules, evaluated at the meta-, superclass-, and fine-class levels. Each value (\%) is reported as mean $\pm$ standard deviation across ten trials.}
\label{tab:schedule_accuracy_all}
\centering
\setlength{\tabcolsep}{3.5pt}
\resizebox{\textwidth}{!}{%
\begin{tabular}{llccc}
\toprule
&
& \multicolumn{1}{c}{Meta Accuracy ($N_{\mathrm{class}}=5$)}
& \multicolumn{1}{c}{Super Accuracy ($N_{\mathrm{class}}=20$)}
& \multicolumn{1}{c}{Fine Accuracy ($N_{\mathrm{class}}=100$)} \\
\cmidrule(lr){3-3} \cmidrule(lr){4-4} \cmidrule(lr){5-5}
data & method & [\%] & [\%] & [\%] \\
\midrule

\multirow{4}{*}{10\%}
& Flat     & 69.66$\,\pm\,1.46$ & 50.63$\,\pm\,1.74$ & 36.73$\,\pm\,1.81$ \\
& 2-stage  & 74.04$\,\pm\,1.00$ & 56.21$\,\pm\,1.26$ & 41.42$\,\pm\,1.00$ \\
& 3-stage  & 74.81$\,\pm\,2.02$ & 56.79$\,\pm\,2.57$ & 42.23$\,\pm\,2.47$ \\
& 4-stage  & 77.11$\,\pm\,0.94$ & 59.94$\,\pm\,1.12$ & 45.36$\,\pm\,0.95$ \\
\midrule

\multirow{4}{*}{20\%}
& Flat     & 83.68$\,\pm\,0.35$ & 69.92$\,\pm\,0.55$ & 57.12$\,\pm\,0.68$ \\
& 2-stage  & 84.96$\,\pm\,0.23$ & 72.08$\,\pm\,0.36$ & 58.97$\,\pm\,0.28$ \\
& 3-stage  & 85.59$\,\pm\,0.31$ & 73.08$\,\pm\,0.39$ & 60.15$\,\pm\,0.32$ \\
& 4-stage  & 85.47$\,\pm\,0.27$ & 72.68$\,\pm\,0.32$ & 59.89$\,\pm\,0.29$ \\
\midrule

\multirow{4}{*}{40\%}
& Flat     & 89.43$\,\pm\,0.28$ & 79.75$\,\pm\,0.38$ & 69.03$\,\pm\,0.39$ \\
& 2-stage  & 89.99$\,\pm\,0.11$ & 80.79$\,\pm\,0.25$ & 69.87$\,\pm\,0.36$ \\
& 3-stage  & 90.10$\,\pm\,0.15$ & 80.83$\,\pm\,0.36$ & 70.00$\,\pm\,0.28$ \\
& 4-stage  & 89.70$\,\pm\,0.20$ & 80.20$\,\pm\,0.36$ & 69.38$\,\pm\,0.25$ \\
\midrule

\multirow{4}{*}{100\%}
& Flat     & 93.45$\,\pm\,0.18$ & 87.19$\,\pm\,0.20$ & 78.91$\,\pm\,0.25$ \\
& 2-stage  & 93.58$\,\pm\,0.11$ & 87.38$\,\pm\,0.13$ & 79.03$\,\pm\,0.24$ \\
& 3-stage  & 93.54$\,\pm\,0.11$ & 87.36$\,\pm\,0.22$ & 79.05$\,\pm\,0.14$ \\
& 4-stage  & 93.11$\,\pm\,0.16$ & 86.90$\,\pm\,0.19$ & 78.28$\,\pm\,0.23$ \\
\bottomrule
\end{tabular}%
}
\end{table}

To examine how the granularity of hierarchical scaffolding affects performance, we compared our main 3-stage setting with 2-stage and 4-stage variants. The flat-learning model can be viewed as a 1-stage setting, in which the model is trained only with the fine-level objective throughout. Our main SCALA configuration used a 3-stage schedule. For a controlled comparison, all variants were trained for the same total of 200 epochs, with only the number of stages and the corresponding stage partitions changed (Table~\ref{tab:schedule_accuracy_all}).

For the 2-stage variant, we removed the meta-level stage from the 3-stage hierarchy and trained the model with superclass-level supervision, followed by fine-level supervision. For the 4-stage variant, we introduced an additional coarser stage before the original meta-level stage by splitting the label space into two broad groups: \textit{Living things} and \textit{Non-living things}. The \textit{Living things} group included \textit{Land mammals}, \textit{Plants}, and \textit{Other creatures}, whereas the \textit{Non-living things} group included \textit{Indoor objects} and \textit{Outdoor objects and scenes}. The resulting 4-stage schedule therefore proceeded as living/non-living $\rightarrow$ meta $\rightarrow$ superclass $\rightarrow$ fine. We fixed the onset of fine-level supervision to epoch 41 across all hierarchical settings, so that each model received 40 epochs of coarser-level training before switching to the fine-level objective. Accordingly, the stage switch points were set to epoch 40 for the 2-stage setting, epochs 20 and 40 for the 3-stage setting, and epochs 10, 25, and 40 for the 4-stage setting.


\subsubsection{Stage transition epochs}

\begin{table}[H]
\caption{Performance under different training set sizes and stage-switching schedules in the 3-stage setting, evaluated at the meta-, superclass-, and fine-class levels. Each value (\%) is reported as mean $\pm$ standard deviation across ten trials. Bold indicates current setup of main experiment with scarce training.}
\label{tab:stage_switching_accuracy}
\centering
\setlength{\tabcolsep}{3.5pt}
\resizebox{\textwidth}{!}{%
\begin{tabular}{llccc}
\toprule
&
& \multicolumn{1}{c}{Meta Accuracy ($N_{\mathrm{class}}=5$)}
& \multicolumn{1}{c}{Super Accuracy ($N_{\mathrm{class}}=20$)}
& \multicolumn{1}{c}{Fine Accuracy ($N_{\mathrm{class}}=100$)} \\
\cmidrule(lr){3-3} \cmidrule(lr){4-4} \cmidrule(lr){5-5}
data & schedule & [\%] & [\%] & [\%] \\
\midrule

\multirow{6}{*}{10\%}
& Flat       & 69.66$\,\pm\,1.46$ & 50.63$\,\pm\,1.74$ & 36.73$\,\pm\,1.81$ \\
& 10/10/180  & 72.89$\,\pm\,1.80$ & 54.68$\,\pm\,2.29$ & 40.41$\,\pm\,2.09$ \\
& \textbf{20/20/160} & $\bm{74.81 \pm 2.02}$ & $\bm{56.79 \pm 2.57}$ & $\bm{42.23 \pm 2.47}$ \\
& 30/30/140  & 75.54$\,\pm\,1.89$ & 57.93$\,\pm\,2.33$ & 42.92$\,\pm\,2.24$ \\
& 40/40/120  & 75.61$\,\pm\,1.59$ & 57.83$\,\pm\,2.03$ & 42.46$\,\pm\,2.20$ \\
& 50/50/100  & 75.33$\,\pm\,1.65$ & 57.34$\,\pm\,1.97$ & 41.32$\,\pm\,2.02$ \\
\midrule

\multirow{6}{*}{20\%}
& Flat       & 83.68$\,\pm\,0.35$ & 69.92$\,\pm\,0.55$ & 57.12$\,\pm\,0.68$ \\
& 10/10/180  & 85.40$\,\pm\,0.34$ & 72.58$\,\pm\,0.42$ & 59.90$\,\pm\,0.43$ \\
& 20/20/160  & 85.59$\,\pm\,0.31$ & 73.08$\,\pm\,0.39$ & 60.15$\,\pm\,0.32$ \\
& 30/30/140  & 85.52$\,\pm\,0.39$ & 72.68$\,\pm\,0.33$ & 59.48$\,\pm\,0.29$ \\
& 40/40/120  & 85.12$\,\pm\,0.39$ & 72.12$\,\pm\,0.46$ & 58.55$\,\pm\,0.31$ \\
& 50/50/100  & 84.70$\,\pm\,0.36$ & 71.17$\,\pm\,0.44$ & 57.14$\,\pm\,0.40$ \\
\midrule

\multirow{6}{*}{40\%}
& Flat       & 89.43$\,\pm\,0.28$ & 79.75$\,\pm\,0.38$ & 69.03$\,\pm\,0.39$ \\
& 10/10/180  & 90.03$\,\pm\,0.20$ & 80.70$\,\pm\,0.37$ & 70.09$\,\pm\,0.32$ \\
& 20/20/160  & 90.10$\,\pm\,0.15$ & 80.83$\,\pm\,0.36$ & 70.00$\,\pm\,0.28$ \\
& 30/30/140  & 90.11$\,\pm\,0.19$ & 80.75$\,\pm\,0.33$ & 69.63$\,\pm\,0.27$ \\
& 40/40/120  & 89.89$\,\pm\,0.16$ & 80.30$\,\pm\,0.29$ & 68.97$\,\pm\,0.34$ \\
& 50/50/100  & 89.66$\,\pm\,0.13$ & 79.81$\,\pm\,0.23$ & 68.17$\,\pm\,0.25$ \\
\midrule

\multirow{6}{*}{100\%}
& Flat       & 93.45$\,\pm\,0.18$ & 87.19$\,\pm\,0.20$ & 78.91$\,\pm\,0.25$ \\
& 10/10/180  & 93.69$\,\pm\,0.17$ & 87.47$\,\pm\,0.24$ & 79.08$\,\pm\,0.25$ \\
& 20/20/160  & 93.54$\,\pm\,0.11$ & 87.36$\,\pm\,0.22$ & 79.05$\,\pm\,0.14$ \\
& 30/30/140  & 93.48$\,\pm\,0.09$ & 87.29$\,\pm\,0.25$ & 78.89$\,\pm\,0.27$ \\
& 40/40/120  & 93.46$\,\pm\,0.20$ & 87.32$\,\pm\,0.13$ & 78.67$\,\pm\,0.25$ \\
& 50/50/100  & 93.30$\,\pm\,0.17$ & 86.97$\,\pm\,0.15$ & 78.24$\,\pm\,0.15$ \\
\bottomrule
\end{tabular}%
}
\end{table}

To examine how the timing of stage transitions affects performance, we fixed the 3-stage SCALA hierarchy (meta $\rightarrow$ super $\rightarrow$ fine) and varied only the switching epochs. Here, we kept the same three-level hierarchy and total training length while changing how long the model remained at each supervision level. All models were trained for 200 epochs in total. We evaluated five schedules, corresponding to meta/super/fine training durations of 10/10/180, 20/20/160, 30/30/140, 40/40/120, and 50/50/100 epochs, respectively. The notation $a/b/c$ denotes the number of epochs assigned to the meta-level, superclass-level, and fine-level objectives. The flat-learning model corresponds to training with the fine-level objective alone for all 200 epochs (Table \ref{tab:stage_switching_accuracy}).

\subsection{Semantic grouping drives the benefit}
\label{app:random}

To test whether the benefit of SCALA depends on semantic coherence, we compared the original semantically aligned hierarchy with a random grouping control. The random grouping condition used the same staged schedule, supervision frequency, and optimization procedure as SCALA, but replaced the meaningful meta- and superclass assignments with randomly assigned groups.

\begin{table}[h]
  \caption{Performance under different training set sizes for the SCALA and random grouping training schemes, evaluated at the meta, superclass, and fine levels. Each value (\%) is reported as mean $\pm$ standard deviation across ten trials.}
  \label{tab:c2f_vs_random_accuracy}
  \centering

  \footnotesize
  \setlength{\tabcolsep}{3pt}
  \renewcommand{\arraystretch}{0.95}

  \begin{tabular}{lccc ccc ccc}
    \toprule
    & \multicolumn{3}{c}{Meta Accuracy ($N_{\mathrm{class}}=5$)}
    & \multicolumn{3}{c}{Superclass Accuracy ($N_{\mathrm{class}}=20$)}
    & \multicolumn{3}{c}{Fine Accuracy ($N_{\mathrm{class}}=100$)} \\
    \cmidrule(lr){2-4} \cmidrule(lr){5-7} \cmidrule(lr){8-10}
    data
    & SCALA [\%] & Random [\%] & $\Delta$
    & SCALA [\%] & Random [\%] & $\Delta$
    & SCALA [\%] & Random [\%] & $\Delta$ \\
    \midrule
    10\%
    & 74.8$\pm$2.0 & 52.2$\pm$1.9 & $\blacktriangledown\ 22.6$
    & 56.8$\pm$2.6 & 41.8$\pm$1.8 & $\blacktriangledown\ 15.0$
    & 42.2$\pm$2.5 & 39.0$\pm$1.6 & $\blacktriangledown\ 3.3$ \\

    20\%
    & 85.6$\pm$0.3 & 66.4$\pm$1.3 & $\blacktriangledown\ 19.2$
    & 73.1$\pm$0.4 & 58.9$\pm$1.2 & $\blacktriangledown\ 14.2$
    & 60.2$\pm$0.3 & 56.9$\pm$0.9 & $\blacktriangledown\ 3.2$ \\

    40\%
    & 90.1$\pm$0.2 & 75.5$\pm$0.6 & $\blacktriangledown\ 14.6$
    & 80.8$\pm$0.4 & 70.1$\pm$0.5 & $\blacktriangledown\ 10.7$
    & 70.0$\pm$0.3 & 68.8$\pm$0.4 & $\blacktriangledown\ 1.2$ \\
    
    100\%
    & 93.5$\pm$0.1 & 83.3$\pm$0.5 & $\blacktriangledown\ 10.3$
    & 87.4$\pm$0.2 & 79.6$\pm$0.3 & $\blacktriangledown\ 7.8$
    & 79.1$\pm$0.1 & 78.6$\pm$0.2 & $\blacktriangledown\ 0.4$ \\
    \bottomrule
  \end{tabular}
\end{table}

Random grouping performed consistently worse than SCALA across all training set sizes and evaluation levels (Table~\ref{tab:c2f_vs_random_accuracy}). The gap was especially large at the meta and superclass levels, where random grouping directly disrupts the intermediate objectives. In the 10\% data setting, SCALA outperformed random grouping by 22.63\% at the meta level and 14.96\% at the superclass level. The advantage also remained at the fine-class level, where SCALA exceeded random grouping by 3.27\%.

These results indicate that the benefit of SCALA cannot be explained by staged training alone. If the improvement came mainly from optimizing easier intermediate objectives, random grouping should have produced comparable performance. Instead, the large gap between semantic and random grouping shows that meaningful label structure is a key component of hierarchical scaffolding. Thus, how classes are grouped matters: the hierarchy must preserve semantic relationships for the scaffold to effectively guide learning.

\subsection{Hierarchical scaffolding generalizes across datasets and architectures}
\label{app:generalization_datasets_architectures}

\begin{table}[h!]
\footnotesize
\centering
\setlength{\tabcolsep}{5pt}
\caption{Architecture generalization on CIFAR-100. Results are reported under the 10\% data regime. Each value (\%) is reported as mean $\pm$ standard deviation across ten trials.}
\label{tab:architecture_generalization}
\begin{tabular}{lccc}
\toprule
Model & Flat [\%] & SCALA [\%] & $\Delta$ \\
\midrule
ResNet-18
& 36.7$\,\pm\,1.8$ & 42.2$\,\pm\,2.5$ & $\blacktriangle\ 5.5\pm2.8$ \\
WideResNet-28-10
& 44.9$\,\pm\,0.7$ & 48.0$\,\pm\,0.7$ & $\blacktriangle\ 3.2\pm0.7$ \\
\bottomrule
\end{tabular}
\end{table}

\begin{table}[h!]
\footnotesize
\centering
\setlength{\tabcolsep}{5pt}
\caption{Dataset and architecture generalization. CIFAR-10 results use 4\% of the training set, and tieredImageNet-H results use 10\% of the training set. Each value (\%) is reported as mean $\pm$ standard deviation across ten trials for CIFAR-10 and across five trials for tieredImageNet-H.}
\label{tab:dataset_architecture_generalization}
\begin{tabular}{llccc}
\toprule
Dataset & Model & Flat [\%] & SCALA [\%] & $\Delta$ \\
\midrule
\multirow{2}{*}{CIFAR-10}
& ResNet-18
& 58.4$\,\pm\,2.2$ & 62.0$\,\pm\,2.3$ & $\blacktriangle\ 3.7\pm2.6$ \\
& ResNet-50
& 51.1$\,\pm\,3.4$ & 56.0$\,\pm\,2.2$ & $\blacktriangle\ 4.8\pm4.3$ \\
\midrule
\multirow{2}{*}{tieredImageNet-H}
& ResNet-18
& 31.3$\,\pm\,0.1$ & 32.8$\,\pm\,0.4$ & $\blacktriangle\ 1.3\pm0.5$ \\
& ResNet-50
& 32.3$\,\pm\,0.7$ & 33.9$\,\pm\,1.0$ & $\blacktriangle\ 1.6\pm1.4$ \\
\bottomrule
\end{tabular}
\end{table}

To examine whether the benefit of SCALA extends beyond the main CIFAR-100 ResNet-18 setting, we evaluated two complementary forms of generalization. First, we tested whether the effect persists across backbone architectures on CIFAR-100 by comparing ResNet-18~\cite{he2016deep} and WideResNet-28-10~\cite{zagoruyko2016wide}. Second, we tested whether the framework extends to different datasets and hierarchy structures by applying SCALA to CIFAR-10 and tieredImageNet-H.

For CIFAR-10, we constructed a three-level hierarchy with 2 meta-classes, 5 superclasses, and 10 fine classes. The two meta-classes were \textit{vehicles} and \textit{animals}; the superclass groupings were (\textit{airplane}, \textit{ship}), (\textit{automobile}, \textit{truck}), (\textit{cat}, \textit{dog}), (\textit{deer}, \textit{horse}), and (\textit{bird}, \textit{frog}). We used 4\% of the CIFAR-10 training set, corresponding to 2k images.

For tieredImageNet-H, we used the WordNet-derived hierarchy released with Making Better Mistakes~\cite{ref22}. The full tieredImageNet-H hierarchy is a tree of height 13 covering 608 fine classes. We selected four coarse ancestor levels above the fine labels, where each value denotes the number of edges from a fine class to the corresponding ancestor. This yielded a five-stage schedule with 4, 11, 63, 108, and 608 classes from coarse to fine. Stage transitions were set to epochs 20, 40, 60, and 80. We used 10\% of the tieredImageNet-H training set, corresponding to 42,560 images.

Across architectures and datasets, SCALA improved performance over the corresponding flat-learning model. These results suggest that the benefit of hierarchical scaffolding is not tied to a single backbone or dataset, but can extend to different label spaces and hierarchy structures. At the same time, the magnitude of the improvement varied across settings, indicating that hierarchy construction and dataset structure remain important factors in determining the strength of the effect.



\end{document}